\documentclass[letterpaper]{article} 
\usepackage{aaai2026}  
\usepackage{times}  
\usepackage{helvet}  
\usepackage{courier}  
\usepackage[hyphens]{url}  
\usepackage{graphicx} 
\usepackage{natbib}  
\usepackage{caption} 
\usepackage{algorithm}
\usepackage{algorithmic}

\usepackage{amsmath} 
\usepackage{booktabs} 
\usepackage{tabularx}  
\usepackage{multirow}  
\usepackage{booktabs}  

\usepackage{newfloat}
\usepackage{listings}
\DeclareCaptionStyle{ruled}{labelfont=normalfont,labelsep=colon,strut=off} 
\floatstyle{ruled}
\newfloat{listing}{tb}{lst}{}
\floatname{listing}{Listing}
\usepackage{amsmath} 
\usepackage{multirow}
\usepackage{booktabs}

\title{C\textsuperscript{3}TG: Conflict-aware, Composite, and Collaborative Controlled Text Generation}
\author{
    Yu Li\textsuperscript{\rm 1},
    Zhe Yang\textsuperscript{\rm 2},
    Yi Huang\textsuperscript{\rm 2,3}\thanks{Corresponding author.},
    Xin Liu\textsuperscript{\rm 4},
    Guilin Qi\textsuperscript{\rm 1}
}
\affiliations{
    \textsuperscript{\rm 1}School of Computer Science and Engineering, Southeast University, Nanjing,  China\\
    \textsuperscript{\rm 2}JIUTIAN Research, China Mobile, China\\
    \textsuperscript{\rm 3}Department of Computer Science and Technology, Tsinghua University, China\\
    \textsuperscript{\rm 4}UniSA STEM, University of South Australia, Adelaide,  Australia\\

    \{yuli11, gqi\}@seu.edu.cn, 
    \{yangzhe, huangyi\}@cmjt.chinamobile.com,
    xin.liu@mymail.unisa.edu.au

}

\begin{document}

\maketitle

\begin{abstract}
Recent advancements in large language models (LLMs) have demonstrated remarkable text generation capabilities. However, controlling specific attributes of generated text remains challenging without architectural modifications or extensive fine-tuning. Current methods typically toggle a single, basic attribute but struggle with precise multi-attribute control. In scenarios where attribute requirements conflict, existing methods lack coordination mechanisms, causing interference between desired attributes. Furthermore, these methods fail to incorporate iterative optimization processes in the controlled generation pipeline.
To address these limitations, we propose \textbf{C}onflict-aware, \textbf{C}omposite, and \textbf{C}ollaborative Controlled Text Generation (C\textsuperscript{3}TG), a two-phase framework for fine-grained, multi-dimensional text attribute control. During generation, C\textsuperscript{3}TG selectively pairs the LLM with the required attribute classifiers from the 17 available dimensions and employs weighted KL-divergence to adjust token probabilities. The optimization phase then leverages an energy function combining classifier scores and penalty terms to resolve attribute conflicts through iterative feedback, enabling precise control over multiple dimensions simultaneously while preserving natural text flow.
Experiments show that C\textsuperscript{3}TG significantly outperforms baselines across multiple metrics including attribute accuracy, linguistic fluency, and output diversity, while simultaneously reducing toxicity. These results establish C\textsuperscript{3}TG as an effective and flexible solution for multi-dimensional text attribute control that requires no costly model modifications.
\end{abstract}



\section{Introduction}

Recent advancements in large language models (LLMs) have revolutionized text generation with their remarkable capabilities~\cite{training,table,mateval,harnessing}. However, precisely controlling fine-grained textual attributes—such as emotion, style, or topic—remains challenging without architectural modifications or extensive fine-tuning~\cite{style}. This goal, known as Controlled Text Generation (CTG), demands techniques that can dynamically modulate model outputs while preserving overall text quality. However, the complexity increases substantially when multiple attributes must be satisfied simultaneously~\cite{magic,tara}, as these attributes may exhibit overlapping or conflicting characteristics: adjusting one attribute can attenuate or amplify others~\citep{palette,sf-gen}. Moreover, current controlled text generation frameworks typically lack mechanisms for iterative feedback refinement~\cite{locateedit}.

\begin{figure}
    \centering
    \includegraphics[width=1\columnwidth]{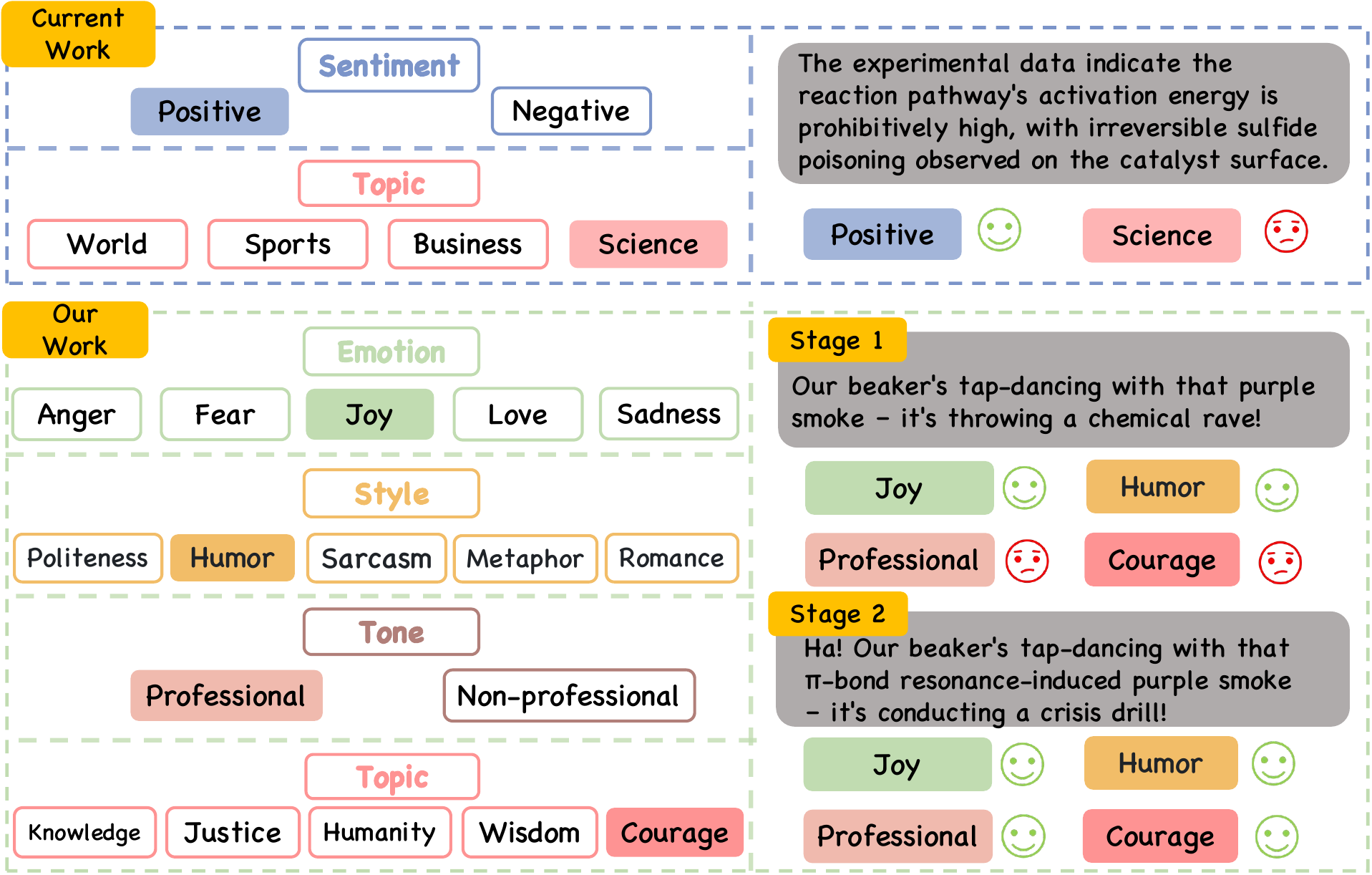}
    \caption{
    Unlike existing methods that control few attributes and handle conflicts poorly, C\textsuperscript{3}TG offers broad attribute control and resolves conflicts effectively.
    \label{motivation}}
\end{figure}

Existing controlled text generation methodologies can be systematically categorized into two principal methods. The first method directly modulates the language model's decoding distribution~\cite{darb,ebdlm}. For instance, PPLM~\cite{pplm} manipulates hidden state gradients during generation to bias text toward target attributes, while GeDi~\cite{gedi} employs a generative discriminator to estimate attribute probabilities for candidate tokens and incorporates this information into the decoding process. Other research has explored latent space energy functions for control, including MacLaSa~\cite{maclasa}, COLD~\cite{cold}, and BOLT~\cite{bolt}, which construct energy functions in latent spaces or over contiguous soft sequences, combining them with gradient sampling or adjustable bias mechanisms to achieve concurrent multi-attribute control while maintaining linguistic fluency~\cite{detoxifying}. 
The second method implements indirect control strategies, such as prompting~\cite{prompt} and fine-tuning~\cite{training,rsacontrol}. The former augments inputs with natural language instructions to guide generation, offering a concise method that remains difficult to calibrate regarding attribute intensity~\cite{prompt-new1}. The latter retrains models on attribute-labeled data, enhancing sensitivity to specific attributes but often requiring substantial computational resources and lacking flexibility~\cite{focused}.
In summary, current methods exhibit several significant limitations~\cite{ctgsurvey}. First, existing methods focus on regulation of individual or simplistic attributes, lacking the capability for fine-grained control over complex multi-attribute combinations. Second, when multiple attributes coexist with potential conflicts, current techniques demonstrate insufficient mechanisms for resolving attribute interference. Finally, these methods fail to support progressive text refinement through iterative optimization processes.

To address these limitations, we propose C\textsuperscript{3}TG (\textbf{C}onflict-aware, \textbf{C}omposite, and \textbf{C}ollaborative Controlled Text Generation), a framework that effectively mitigates attribute conflicts by coordinating an LLM with compact auxiliary models. C\textsuperscript{3}TG pairs a powerful LLM with lightweight BERT-based classifiers in a two-phase method. During generation, it integrates attribute-specific probability distributions via weighted KL divergence terms, ensuring each token selection reflects all target attributes. In the optimization phase, C\textsuperscript{3}TG constructs an energy function combining classifier scores with conflict penalty terms to guide iterative text refinement, progressively improving attribute alignment while maintaining fluency. This collaboration between ``large'' generators and ``small'' evaluators enables flexible, fine-grained, conflict-aware control without costly retraining or architectural modifications. 
Extensive story‑generation experiments show that C\textsuperscript{3}TG preserves fluency and diversity while controlling multiple attributes. Additional evaluations on toxicity datasets show significant reductions in harmful content generation. Moreover, our dedicated ``opposites and conflicts'' experiments highlight C\textsuperscript{3}TG's superior capacity to handle overlapping and conflicting attribute requirements. Our contributions are summarized as follows:

\begin{itemize}
    \item We propose C\textsuperscript{3}TG, pairing LLMs with specialized BERT classifiers across 17 subcategories of emotion, style, tone, and topic, significantly expanding attribute control capabilities while mitigating toxic content generation.
    \item We integrate a weighted KL divergence for attribute distribution fusion during generation and implement a composite energy function (classifier scores plus conflict penalties) during optimization, enabling flexible real-time multi-attribute control that outperforms existing static methods in both stability and toxicity suppression.
    \item Empirical evaluations on ROCStories and WritingPrompts datasets demonstrate C\textsuperscript{3}TG's superiority over mainstream baselines in attribute accuracy, fluency, 
    and diversity—findings further validated through comprehensive human evaluation.
\end{itemize}

\section{Related Work}
\subsection{Controlled Text Generation via Decoding}
Recent research has advanced CTG by manipulating decoding distributions \cite{control3}. These methods include: gradient-based methods like PPLM \cite{pplm}, COLD \cite{cold}, and BOLT \cite{bolt} that adjust hidden states during inference; latent-space techniques such as MacLaSa \cite{maclasa} and LatentOps \cite{LatentOps} operating in continuous attribute spaces \cite{sld}; constraint frameworks like MUCOCO \cite{MUCOCO} and PriorControl \cite{PriorControl}; and score-mixing methods including Mix\&Match \cite{mix} and Palette \cite{palette}. While these methods preserve the base model and enable fine-grained control, their heavy‑handed decoding interventions add implementation complexity, frequently undermine fluency and coherence, and—crucially—provide no feedback‑driven refinement, a gap that becomes acute when resolving conflicts in multi‑attribute scenarios~\cite{airdecoding}.

\subsection{Controlled Text Generation via Indirect Strategies}
Indirect strategies guide pretrained models without internal modifications. Prompt-based methods insert control signals, offering parameter-free solutions but providing only coarse control and limited efficacy with conflicting attributes \cite{prompt1,prompt2}. Fine-tuning methods—including prefix-based adaptation \cite{prefix}, ProSwitch \cite{proswitch}, and prompt tuning \cite{training3}—enable precise control but require substantial annotation and risk overfitting \cite{pilm}. Plug-and-play frameworks such as DATG \cite{Datg} and LiFi \cite{lifi} achieve lightweight control without modifying the base LLM, but both depend on quality labeled data to train their attribute classifiers \cite{crlctg}. Despite methodological diversity, these methods universally struggle with two fundamental challenges: resolving multi-attribute conflicts and enabling feedback-driven refinement to enhance attribute alignment without compromising coherence \cite{scope}.

\begin{figure*}
    \centering
    \includegraphics[width=0.70\textwidth]{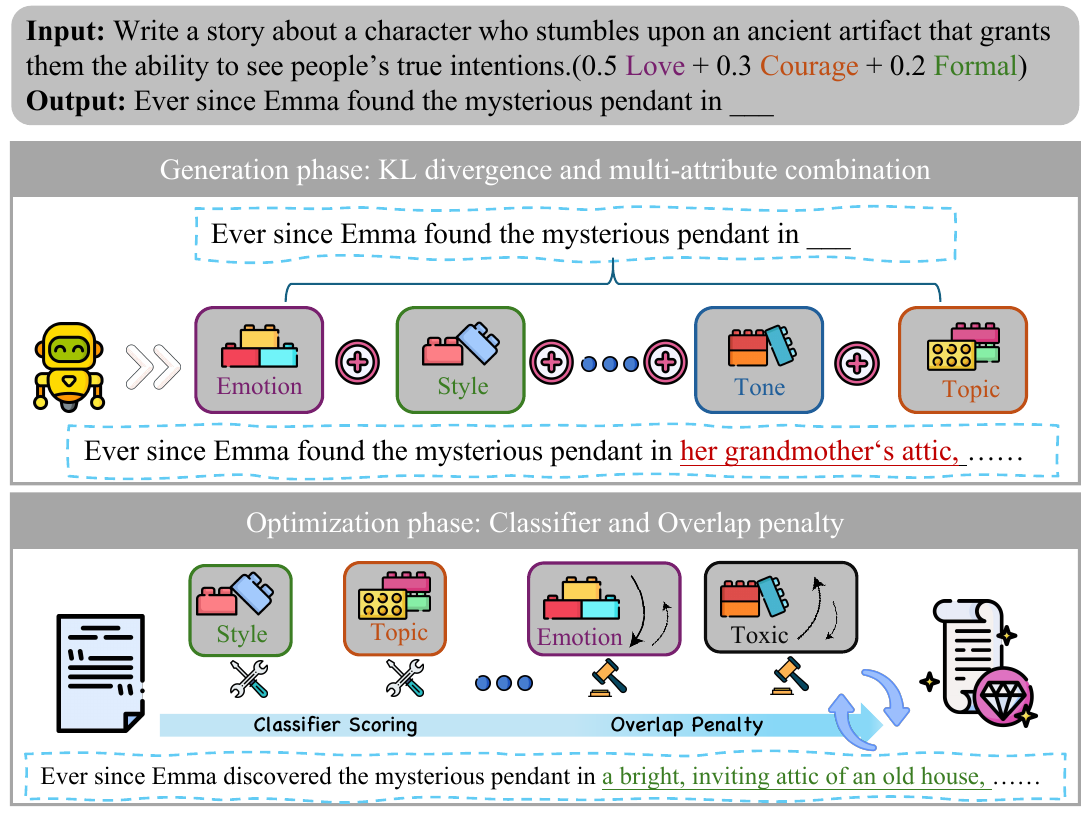}
    \caption{Flowchart of the overall process of the C\textsuperscript{3}TG framework, including the generation phase and the optimization phase, leading to the generation of a text that meets the user's requirements.\label{framework}}
\end{figure*}

\section{Problem Formulation}
Let \(\mathcal{V}\) denote the vocabulary and \(x=(x_1,\dots,x_T)\in\mathcal{V}^T\) represent a sequence of tokens. We consider the controlled text generation problem over an attribute set \(\mathcal{A}=\{A_1,\dots,A_n\}\) with corresponding target intensities \(T_i\in[0,1]\). Formally, we aim to solve the constrained optimization problem:

\begin{equation}\label{eq:opt}
\begin{aligned}
\max_{x}\;&\mathrm{Fluency}(x)  \\
\text{s.t.}\;&C_{A_i}(x)\approx T_i,\quad \forall\,i\in\{1,\dots,n\}.
\end{aligned}
\end{equation}

where \(C_{A_i}(x)\) quantifies the intensity of attribute \(A_i\) in sequence \(x\).

Our framework accepts as input: (1) an initial context \(x_{\mathrm{init}}\) provided by the user, and (2) a set of desired attribute-intensity pairs \(\{(A_i, T_i)\}_{i=1}^n\), such as \textit{Emotion: Joy (0.9), \textit{Style}: Humor (0.85), \textit{Tone}: Professional (0.9), and \textit{Topic}: Courage (0.9)}. The framework outputs an optimized text sequence \(x_{\mathrm{final}}\) that maximizes fluency while satisfying the attribute constraints \(C_{A_i}(x_{\mathrm{final}})\approx T_i\) for all target attributes. Additionally, the system provides real-time attribute intensity estimates \(\{C_{A_i}(x_t)\}_{i=1}^n\) throughout the generation process and guarantees monotonic reduction in toxicity metrics.

\section{Methodology}
\label{methodology}

The overall workflow of C\textsuperscript{3}TG is shown in Figure \ref{framework}. Our framework consists of two principal phases: generation and optimization. During the generation phase, we leverage the foundational language model and incorporate weighted Kullback-Leibler divergence to integrate probability distributions from attribute-specific models corresponding to user-defined targets (e.g., emotion, style, tone, topic). This integration mechanism imposes a differentiable constraint on token selection probabilities, yielding an initial text output that incorporates the specified attribute characteristics.

However, a single-pass generation process frequently proves insufficient for simultaneously satisfying multiple attribute objectives—particularly when attributes exhibit conflicting or interdependent relationships. Consequently, in the optimization phase, we employ discriminative classifiers to quantitatively assess attribute alignment and introduce penalty functions that target attribute conflicts. We formulate a composite energy function combining classifier scores and conflict penalty terms. When attributes deviate from targets or interfere with other attributes after the generation phase, a \textbf{Feedback Agent} generates rewriting prompts based on the differential scores from classifiers and penalty terms. This agent then guides the model through a three-stage rewriting process to produce the final text. The resulting closed-loop system transitions from initial generation to structured refinement, establishing an optimal balance between attribute target satisfaction and linguistic coherence.

\subsection{Generation Phase}\label{generation}
We adopt Llama2 \cite{llama} as the base language model for generation.
In parallel, we fine-tune independent Llama2 models on attribute-specific corpora, obtaining $n$ attribute models—emotion, style, tone, topic, and toxicity—each yielding a prior distribution $Q_i(\cdot\mid x_{1:t-1})$.
\paragraph{Optimization Objective:}
Given user-specified importance scores $\{\lambda_i\}_{i=1}^{n}$ ($\lambda_i \ge 0$)
(e.g., \textit{Joy} 0.9, \textit{Polite} 0.8), we seek a distribution
$P(\cdot\mid x_{1:t-1})$ that balances all attributes by minimizing the
weighted KL divergence:

\begin{align}
\label{eq:obj}
\mathcal{J}[P] \;=\;&
  \sum_{i=1}^{n} \lambda_i\,
    D_{\mathrm{KL}}\!\bigl(
      P(\,\cdot\mid x_{1:t-1}) \,\big\|\, Q_i(\,\cdot\mid x_{1:t-1})
    \bigr), \nonumber\\[4pt]
&\text{s.t. } \sum_{x \in \mathcal{V}} P(x \mid x_{1:t-1}) = 1.
\end{align}

\paragraph{Solution:} By applying Lagrange multipliers, we derive the optimal distribution:

\begin{align}
\label{eq:p_star}
P^{*}(x \mid x_{1:t-1})
  &= \frac{\displaystyle
        \prod_{i=1}^{n}
        Q_i(x \mid x_{1:t-1})^{\lambda_i/\Lambda}}
       {\displaystyle
        \sum_{x' \in \mathcal V}
        \prod_{i=1}^{n}
        Q_i(x' \mid x_{1:t-1})^{\lambda_i/\Lambda}}, \nonumber\\[4pt]
\Lambda &= \sum_{i=1}^{n} \lambda_i > 0.
\end{align}

Thus, we sample tokens from the \emph{weighted geometric mean} of attribute priors, with $\lambda_i/\Lambda$ controlling attribute $A_i$'s influence. 
Unsatisfactory outputs enter the optimization phase for rewriting.

\subsection{Optimization Phase}

The initial generation frequently produces text that deviates from target attribute intensities or exhibits inter-attribute conflicts, where enhancing one attribute may inadvertently suppress or amplify others. To address these challenges, we formulate an energy-based optimization framework that integrates BERT~\cite{bert} classifier feedback with specialized penalty terms to mitigate dimensional conflicts during iterative refinement.

\subsubsection{Attribute Classifier Scores: }

In multi-attribute controlled generation, we decompose each high-level attribute into fine-grained dimensions. For instance, the emotion attribute encompasses sub-categories such as joy, sadness, and love. Let ${A_1, A_2, \ldots, A_n}$ denote the set of controllable dimensions, with $C_{A_i}(x)$ representing the BERT-based classifier that quantifies dimension $A_i$ given text $x$.

Ideally, $C_{A_i}(x)$ should approximate the specified target value $T_i$. We measure the aggregate deviation through:
\begin{equation}
E_{\text{classify}}(x)=\sum_{i=1}^{n}
\alpha_i\lvert C_{A_i}(x)-T_i\rvert,
\label{eq:classify_term}
\end{equation}
where $\alpha_i$ denotes the importance weight of dimension $A_i$, initially specified by the user.
After computing Eq.~\eqref{eq:classify_term}, we store each score $C_{A_i}(x)$ and its deviation from the target, enabling adaptive prioritization of dimensions with large discrepancies in later iterations. These deviation metrics feed the \textit{Feedback Agent} (Llama2) described later, which turns them into targeted rewriting directives.

\subsubsection{Dimensional Stability Penalties: }

During optimization, purely pursuing target dimensional improvements without safeguarding related dimensions frequently induces unintended interference effects across the attribute space. To address this challenge, we formulate dimensional stability penalties that quantify and constrain perturbations in non-primary dimensions during each optimization iteration.

Let $x_{\mathrm{prev}}$ denote the text before rewriting and $x$ represent the current iteration result. While each iteration focuses on specific ``primary optimization dimensions'', we designate the remaining $k$ dimensions as ``stability-constrained dimensions'' ${A_1, A_2, \dots, A_k}$. For instance, when optimizing ``Emotion-Joy'', both other dimensions within the same attribute (such as ``Sadness'', ``Love'') and dimensions from different attributes (such as ``Style-Formal'', ``Topic-Courage'') collectively form the stability-constrained set. To quantify dimensional fluctuations during rewriting, we define the penalty function:
\begin{equation}
\Omega_{\text{overlap}}(x) 
= \sum_{i=1}^{k} \beta_i \,\bigl\lvert C_{A_i}(x) - C_{A_i}(x_{\text{prev}}) \bigr\rvert,
\label{eq:overlap_term}
\end{equation}

where $\beta_i$ represents the penalty coefficient for dimension $A_i$. Due to inter-dimensional correlations, we assign dimension-specific penalties $\beta_i$ based on the correlations between attributes. The function $C_{A_i}(\cdot)$ represents the classifier's score for dimension $A_i$, with the absolute difference constraining both enhancement and suppression effects.

To facilitate precise multi-stage refinement, we focus on a few dimensions in each iteration while aggregating all non-targeted dimensions into the stability-constrained set ${A_1,\dots,A_k}$ governed by Eq.~\eqref{eq:overlap_term}. For example, when optimizing ``Humor'', other style dimensions along with all emotion, topic, and tone attributes are incorporated into the stability-constrained set. As refinement progresses, this constrained set evolves dynamically to maintain global multi-dimensional equilibrium.
At the end of each round, the \textbf{Feedback Agent} records both the aggregate penalty $\Omega_{\mathrm{overlap}}(x)$ and individual dimensional shifts $\bigl|C_{A_i}(x)-C_{A_i}(x_{\mathrm{prev}})\bigr|$. 
Together with the primary‑dimension classifier scores, these metrics supply the information the agent needs to craft the next prompt, allowing fine‑grained tuning while preserving attribute balance.

\begin{figure}
    \centering
    \includegraphics[width=1\columnwidth]{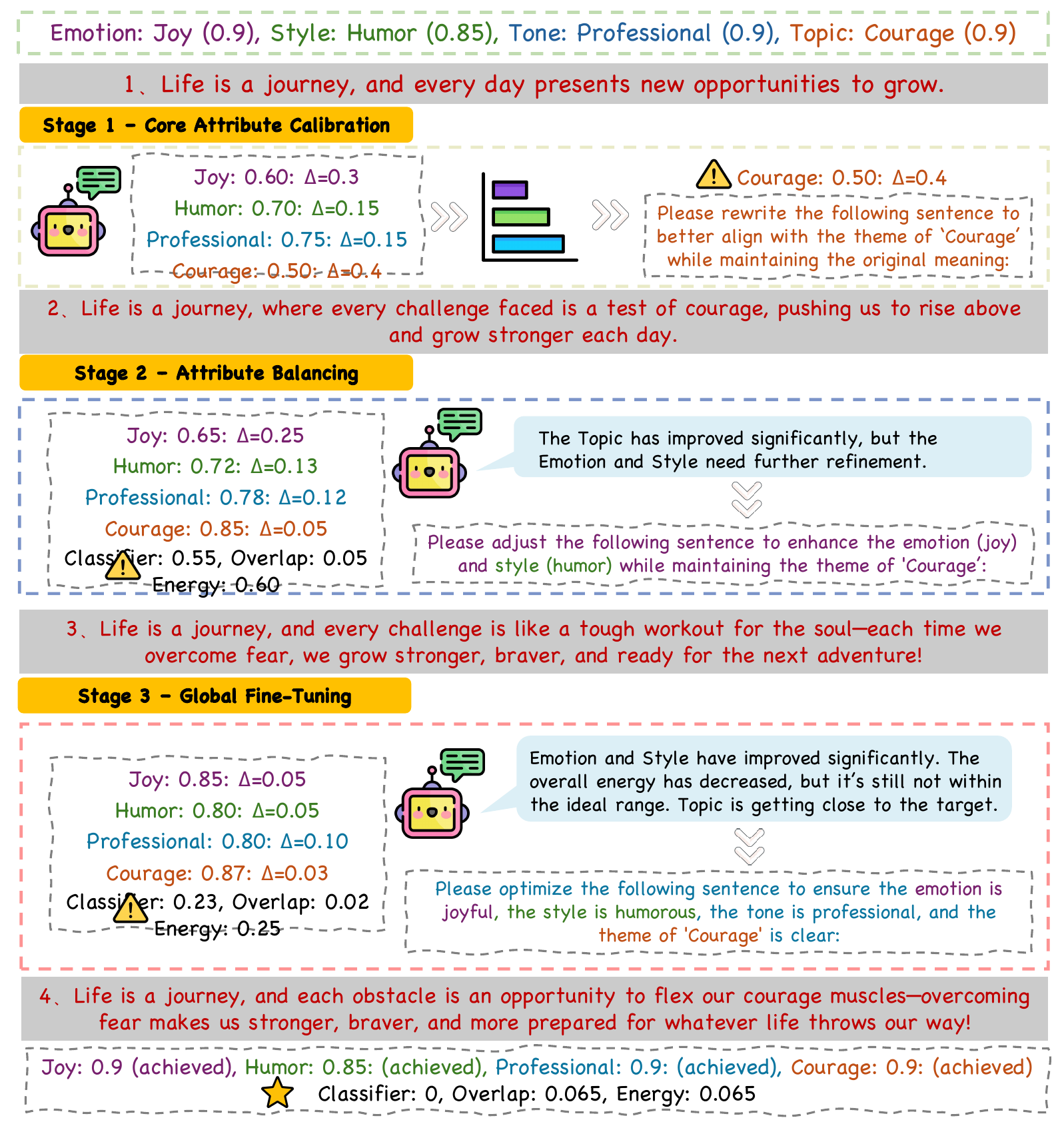}
    \caption{Schematic diagram of the complete Chain-of-Prompt workflow in the optimization phase of the C\textsuperscript{3}TG framework.\label{opt}}
\end{figure}

\subsubsection{Energy Function: }
\label{energy-function}

After obtaining the classifier scores (Eq. \eqref{eq:classify_term}) with the penalty term (Eq. \eqref{eq:overlap_term}), our energy function is:


\begin{equation}\label{eq:final_energy}
\begin{aligned}
E(x)=\;&\sum_{i=1}^{n} \alpha_i\,\bigl|C_{A_i}(x)-T_i\bigr|\\[2pt]
      &+ \sum_{j=1}^{k} \beta_j\,\bigl|C_{A_j}(x)-C_{A_j}(x_{\mathrm{prev}})\bigr|.
\end{aligned}
\end{equation}

where the first term measures alignment with target values across all optimization dimensions, while the second term constrains perturbations in stability-constrained dimensions. The weight $\alpha_i$ is user-specified to reflect dimension importance, while $\beta_j$ is determined based on experimentally derived attribute correlations.

Leveraging the individual terms of the energy function and their step-to-step changes, a zero-shot Llama2-7B \textit{Feedback Agent} turns this data into rewrite prompts and drives a three-phase iterative optimization loop:

\begin{enumerate}
\item Evaluates the energy function (Eq. \eqref{eq:final_energy}) and records deviations $\Delta_i=\lvert C_{A_i}(x)-T_i\rvert$ for each iteration;
\item Constructs a priority-ordered correction queue based on deviation magnitudes, focusing attention on dimensions with largest target-value discrepancies;
\item Synthesizes precise rewriting prompts that explicitly specify dimensional adjustments (e.g., ``Increase dimension $A_j$, slightly reduce $A_k$, maintain $A_m$'');
\item Monitors convergence criteria and terminates optimization when energy values satisfy predefined thresholds.
\end{enumerate}

The agent refines target attributes through a three‑stage optimization, while safeguarding the remaining dimensions from undue interference. This method efficiently navigates the multi-dimensional attribute space until reaching convergence criteria or selecting the text with minimal energy value after the maximum allowed iterations.

\subsubsection{Agent-Guided Chain-of-Prompt Refinement:}

To jointly control multiple attributes, avert conflicts, and preserve fluency, we devise a three‑stage optimization procedure:

\textbf{Stage 1 - Core Attribute Calibration:}  
  At the beginning of each iteration, we aggregate classifier scores and penalty terms, with the \textit{Feedback Agent} evaluating dimensional deviations $\bigl|C_{A_i}(x)-T_i\bigr|$. The Agent prioritizes dimensions requiring significant adjustment and generates targeted prompts to efficiently minimize the global energy function. For instance: ``\emph{Please rewrite the text to better align with the theme of "Courage" while preserving semantic content: $\langle$Original text$\rangle$}''.

\textbf{Stage 2-Attribute Balancing Adjustment:}  
  After core calibration, we count the change in scores for all non-optimized dimensions  
  \(\Delta_i = \bigl|C_{A_i}(x)-C_{A_i}(x_{\mathrm{prev}})\bigr|\) and reassess the bias  
  \(\bigl|C_{A_i}(x)-T_i\bigr|\). The \textit{Feedback Agent} then generates precision-calibrated prompts incorporating intensity modifiers ("slightly," "significantly") for dimensions with persistent deviations, while explicitly specifying attributes requiring stability to prevent emergent conflicts: ``\emph{Please modify the text to significantly enhance joy and humor while maintaining the theme of "Courage": $\langle$Phase 1 output$\rangle$}''.

\textbf{Phase 3 - Global fine-tuning:}  
After the second rewrite, the agent craft a single, consolidated prompt that uses the composite energy $E(x)$ and the latest attribute metrics to direct final, per-attribute fine-tuning—e.g., “\emph{Please polish the text to keep a pleasant tone, slightly raise humor and formality, and retain the clear ‘Courage’ theme: $\langle$Phase 2 output$\rangle$}.” 
This third stage performs the last adjustments to bring every attribute as close as possible to its target score, driving $E(x)$ below the convergence threshold $\tau$(\(\tau=0.025\)).

In summary, the following steps are performed for each iteration round:  
\begin{enumerate}
\item The \textit{Feedback Agent} formulates a context-aware prompt based on current classifier and penalty evaluations;
\item The LLM generates text $x_{\mathrm{new}}$ conditioned on the prompt;
\item We compute the energy differential $\Delta E = E(x_{\mathrm{new}})-E(x_{\mathrm{prev}})$ and terminate if both $\Delta E<0$ and $E(x_{\mathrm{new}})\leq\tau$; otherwise, we proceed to the next phase or iteration.
\end{enumerate}

If convergence criteria remain unsatisfied after reaching the maximum iteration limit, we return the lowest-energy text. Figure \ref{opt} summarizes the three-phase optimization.

\section{Experiment}
\label{experiment}

\begin{table*}[t]
\centering

\small

\begin{tabular}{l ccc ccc c}
\toprule
\multirow{2}{*}{\textbf{Method}}
& \multicolumn{3}{c}{\textbf{ROC}} & \multicolumn{3}{c}{\textbf{WP}} & \multirow{2}{*}{\textbf{Tox.}$\downarrow$} \\
\cmidrule(lr){2-4}\cmidrule(lr){5-7}
& \textbf{Acc.}$\uparrow$ & \textbf{PPL}$\downarrow$ & \textbf{Dist-1/2/3}$\uparrow$
& \textbf{Acc.}$\uparrow$ & \textbf{PPL}$\downarrow$ & \textbf{Dist-1/2/3}$\uparrow$ &  \\
\midrule
\multicolumn{8}{l}{\emph{Controllable text generation by controlling decoding distribution}} \\
COLD             & 24.35 & 21.07 & 0.08/0.10/0.22 & 20.50 & 24.54 & 0.06/0.08/0.18 & 0.53 \\
BOLT             & 36.54 & 17.33 & 0.09/0.28/0.38 & 32.07 & 20.52 & 0.08/0.26/0.36 & 0.76 \\
MuCola           & 27.93 & 16.89 & 0.13/0.24/0.33 & 25.12 & 19.83 & 0.10/0.21/0.30 & 0.58 \\
MacLaSa          & 31.54 & 16.35 & 0.11/0.25/0.33 & 28.45 & 19.01 & 0.08/0.22/0.30 & 0.60 \\
PriorControl     & 22.56 & 18.54 & 0.09/0.25/0.29 & 19.36 & 22.07 & 0.07/0.22/0.25 & 0.66 \\
LatentOps        & 33.55 & 14.57 & 0.18/0.22/0.31 & 30.00 & 17.39 & 0.13/0.17/0.26 & 0.55 \\
PPLM             & 32.39 & 15.04 & 0.19/0.23/0.39 & 29.74 & 18.20 & 0.16/0.20/0.36 & 0.39 \\
Mix\&Match       & 49.78 & 22.71 & 0.21/0.24/0.37 & 45.00 & 26.46 & 0.11/0.20/0.32 & 0.47 \\
Model Arithmetic & 87.53 & 11.08 & 0.37/0.68/0.81 & 84.23 & 14.30 & 0.33/0.50/0.75 & 0.16 \\
\midrule
\multicolumn{8}{l}{\emph{Controllable text generation by indirect control strategy}} \\
LLM-based Prompt      & 89.45 & 5.37  & 0.47/0.71/0.89 & 80.02 & 9.65  & 0.42/0.50/0.82 & 0.29 \\
LLM-based Fine-tuning & 79.03 & 8.53  & 0.39/0.68/0.84 & 75.00 & 10.50 & 0.38/0.54/0.80 & 0.16 \\
\midrule
\textbf{C\textsuperscript{3}TG (Ours)}
& \textbf{90.39} & \textbf{4.04} & \textbf{0.53}/\textbf{0.74}/\textbf{0.90}
& \textbf{85.56} & \textbf{3.68} & \textbf{0.47}/\textbf{0.55}/\textbf{0.84}
& \textbf{0.12} \\
\bottomrule
\end{tabular}
\caption{Comparison of automated evaluation results for C\textsuperscript{3}TG and baseline methods on the ROC and WP datasets.}
\label{automated_evaluation}
\end{table*}

\begin{table*}[tb]
\centering

\small
\begin{tabular}{l ccc ccc}
\toprule
\multirow{2}{*}{\textbf{Method}}
& \multicolumn{3}{c}{\textbf{ROC}} 
& \multicolumn{3}{c}{\textbf{WP}} \\
\cmidrule(lr){2-4}\cmidrule(lr){5-7}

& \textbf{Topic\%}$\uparrow$ 
& \textbf{Fluency}$\uparrow$ 
& \textbf{Diversity}$\uparrow$
& \textbf{Topic\%}$\uparrow$ 
& \textbf{Fluency}$\uparrow$ 
& \textbf{Diversity}$\uparrow$ \\
\midrule
\multicolumn{7}{l}{\emph{Controllable text generation by controlling decoding distribution}} \\
COLD             
& 1.07 & 1.68 & 1.32 
& 0.98 & 1.55 & 1.26 \\
BOLT             
& 2.89 & 2.34 & 2.76 
& 2.45 & 1.98 & 2.35 \\
MuCola           
& 2.71 & 2.62 & 2.52 
& 2.38 & 2.33 & 2.20 \\
MacLaSa          
& 2.95 & 2.84 & 2.64 
& 2.61 & 2.48 & 2.31 \\
PriorControl     
& 2.30 & 2.12 & 2.05 
& 2.16 & 1.95 & 1.90 \\
LatentOps        
& 3.12 & 2.92 & 2.84 
& 2.85 & 2.65 & 2.55 \\
PPLM             
& 3.40 & 2.53 & 3.06 
& 3.05 & 2.23 & 2.74 \\
Mix\&Match       
& 3.86 & 3.11 & 3.54 
& 3.52 & 2.76 & 3.23 \\
Model Arithmetic 
& 4.20 & 3.85 & 4.07 
& 3.77 & 3.47 & 3.69 \\
\midrule
\multicolumn{7}{l}{\emph{Controllable text generation by indirect control strategy}} \\
LLM-based Prompt     
& 4.73 & 3.97 & 4.03 
& 3.24 & 3.18 & 3.76 \\
LLM-based Fine-tuning
& 4.33 & 3.82 & 4.28 
& \textbf{3.65} & 3.57 & 3.42 \\
\midrule
\textbf{C\textsuperscript{3}TG (Ours)} 
& \textbf{4.74} & \textbf{4.53} & \textbf{4.45 }
& \textbf{3.65} & \textbf{3.88} & \textbf{4.05} \\
\bottomrule
\end{tabular}
\caption{Comparison of human evaluation results for C\textsuperscript{3}TG and baseline methods on the ROC and WP datasets.}
\label{human_evaluation}
\end{table*}

We conduct a comprehensive empirical evaluation of C\textsuperscript{3}TG to address the following research questions:

\begin{itemize}
    \item \textbf{RQ1:} How does C\textsuperscript{3}TG compare to state-of-the-art methods across automated metrics and human evaluations?
    \item \textbf{RQ2:} How effectively does C\textsuperscript{3}TG handle attribute conflicts and interactions?
    \item \textbf{RQ3:} What is the contribution of each component in C\textsuperscript{3}TG to system effectiveness?
\end{itemize}

\subsection{Experimental setup}

\paragraph{Dataset}
We select two English story scenarios: ROCStories(ROC) and WritingPrompts(WP)\cite{roc}. ROC emphasizes causal and temporal relationships in everyday scenarios, while WP provides structurally rich prompt-story pairs. This combination enables rigorous assessment of multi-attribute control and our iterative optimization framework across diverse textual contexts.

\paragraph{Control Dimensions}
C\textsuperscript{3}TG controls five primary attribute categories: (1) Emotion (anger, fear, joy, love, sadness, surprise); (2) Style (politeness, romantic, humor, sarcasm, metaphorical); (3) Tone (professional, casual); (4) Topic (knowledge, justice, humanity, courage); and (5) Toxicity (toxic, non-toxic). For classifier training, we utilize publicly available datasets: Social Network Sentiment corpus, xSLUE style annotation collection, Domain Q\&A and Instructions repository, and Toxicity Review dataset.

\paragraph{Baselines}
We compare C\textsuperscript{3}TG with two main categories of mainstream methods: methods that directly intervene in the decoding distribution, including COLD \cite{cold}, BOLT \cite{bolt}, MuCoLa \cite{MuCola}, MacLaSa \cite{maclasa}, PriorControl \cite{ PriorControl}, LatentOps \cite{LatentOps}, PPLM \cite{pplm}, Mix\&Match \cite{mix}, and Model Arithmetic \cite{modelarith}; the other category is the indirect control strategies including LLM-based Prompt and LLM-based Fine-tuning.

\paragraph{Evaluation Metrics}
For \textbf{automated evaluation}, we employ classifier-based accuracy to measure attribute control precision, language model perplexity to assess textual coherence, Distinct-$n$ metrics to quantify lexical diversity, and toxicity probability via an external API to verify content safety. 
In the \textbf{human evaluation}, five independent domain experts assess system outputs on a 5-point Likert scale across three dimensions—attribute alignment, linguistic fluency, and content diversity—with the final score equal to the average rating.

\paragraph{Time optimization}
We batch the gradients of the user-specified attributes into a single GPU kernel and reuse cached hidden states, so each generation step requires only one forward–backward pass. An energy-based early-stopping criterion trims decoding time by about 40\%. Overall, C³TG is $\mathbf{1.6\times}$ slower than the fastest baseline but still delivers roughly 3\% higher attribute accuracy and 25\% lower toxicity.

\begin{table*}[t]
\centering

\small
\begin{tabular}{l ccc ccc}
\toprule
\multirow{2}{*}{\textbf{Method}}
& \multicolumn{3}{c}{\textbf{Conflict Experiment}} 
& \multicolumn{3}{c}{\textbf{Overlap Experiment}} \\
\cmidrule(lr){2-4}\cmidrule(lr){5-7}
& Average$\downarrow$ & PPL$\downarrow$ & Drift$\downarrow$
& Average$\downarrow$ & PPL$\downarrow$ & Drift$\downarrow$\\
\midrule
Model Arithmetic  & 0.27&  10.48 &  0.38 &  0.22 &  11.03 & 0.31   \\
LLM-based Prompt        & 0.19 &  5.75&  0.25 &  0.12 &  4.96 &  0.29  \\
\textbf{C\textsuperscript{3}TG} & \textbf{0.08} &  \textbf{4.54}&  \textbf{0.16} &  \textbf{0.07}&  \textbf{4.13} & \textbf{0.18}  \\
\bottomrule
\end{tabular}
\caption{Performance of the C\textsuperscript{3}TG in conflict and overlap experiments on the ROC dataset.}
\label{tab:tone-results}
\end{table*}

\begin{table*}[t]
\centering

\small
\begin{tabular}{lcccc cccc}
\toprule
\multirow{2}{*}{\textbf{Method}} 
& \multicolumn{4}{c}{\textbf{ROC}} 
& \multicolumn{4}{c}{\textbf{WP}} \\
\cmidrule(lr){2-5}\cmidrule(lr){6-9}
& \textbf{Acc.}$\uparrow$ & \textbf{PPL}$\downarrow$ & \textbf{Dist-2}$\uparrow$ & \textbf{Tox.}$\downarrow$
& \textbf{Acc.}$\uparrow$ & \textbf{PPL}$\downarrow$ & \textbf{Dist-2}$\uparrow$ & \textbf{Tox.}$\downarrow$ \\
\midrule
\multicolumn{9}{l}{\emph{Components}} \\
w/o Optimization        & 65.22 & 10.53 & 0.35 & 0.38 & 62.17 & 12.64 & 0.29 & 0.42 \\
w/o Generation        & 59.40 & 25.62 & 0.28 & 0.32 & 55.08 & 28.03 & 0.25 & 0.45 \\
w/o Overlap             & 78.46 & 5.11  & 0.47 & 0.36 & 73.56 & 6.88  & 0.41 & 0.43 \\
\midrule
\multicolumn{9}{l}{\emph{Iterations}} \\
1-Iteration & 74.21 & 6.31 & 0.43 & 0.19 & 70.09 & 7.94 & 0.39 & 0.27 \\
2-Iteration & 85.62 & 4.89 & 0.48 & 0.12 & 81.13 & 5.81 & 0.44 & 0.26 \\
\midrule
\textbf{C\textsuperscript{3}TG} 
& \textbf{90.39} & \textbf{4.04} & \textbf{0.74} & \textbf{0.12} 
& \textbf{85.56} & \textbf{3.68} & \textbf{0.55} & \textbf{0.24} \\
\bottomrule
\end{tabular}
\caption{Ablation results of the C\textsuperscript{3}TG framework on ROC and WP datasets. }
\label{ablation}
\end{table*}

\subsection{Overall Experimental Results(RQ1)}

As shown in Tables \ref{automated_evaluation} and \ref{human_evaluation}, which include direct decoding intervention methods as well as methods with indirect control strategies, the C\textsuperscript{3}TG results are listed at the bottom. 
All baselines follow their original specs; single-attribute models received minimal, architecture-preserving tweaks for multi-attribute control, were tuned to their best configuration, and evaluated with identical metrics for fairness.

\paragraph{Automated evaluation}  
Quantitative analysis reveals that C\textsuperscript{3}TG achieves superior balance across accuracy, perplexity, and diversity metrics. To evaluate toxicity mitigation capabilities, we employed the /pol/ dataset \cite{duxing} for controlled text rewriting through the C\textsuperscript{3}TG framework. Toxicity evaluation via an external API service demonstrates that C\textsuperscript{3}TG show C\textsuperscript{3}TG yields the lowest toxicity, confirming that its iterative energy minimization jointly boosts attribute accuracy, fluency, diversity, and suppresses harmful content.

\paragraph{Human evaluation}  
Human evaluation shows C\textsuperscript{3}TG’s significant advantage over all baselines in attribute alignment, fluency, and diversity. Its composite energy function and multi-stage prompt chain jointly reinforce target attributes, resolve conflicts, and preserve related dimensions, producing content that retains natural variation and coherence.

\subsection{Conflict and Overlap Experiment(RQ2)}

To evaluate C\textsuperscript{3}TG's robustness under attribute conflicts, we test it against \textit{Model Arithmetic} and an \textit{LLM-based prompt} baseline on 30\% of ROCStories. We constructed a negative pair (``fear 0.7 vs. joy 1.0'') and a positive pair (``romance 0.7 + love 0.7''), keeping other attributes constant. Performance was measured via Average Absolute Bias, Perplexity (PPL), and Uncontrolled Dimensional Drift (Drift measures the average absolute change in all non-target attribute scores, capturing unintended side-effects).

As shown in Table \ref{tab:tone-results}, in both scenarios, C\textsuperscript{3}TG demonstrates superior performance—achieving minimal bias (0.08/0.07) while maintaining optimal PPL and Drift metrics. Model Arithmetic amplifies positively correlated attributes and prompt-based tuning boosts fluency, but neither approach succeeds in curbing attribute drift. These results confirm that C\textsuperscript{3}TG's integration of classifier feedback with iterative penalty terms delivers superior attribute stability across conflicting and overlapping conditions.

\subsection{Ablation Experiment(RQ3)}


We conduct ablation experiments on ROC and WP datasets to assess component contributions. Four configurations are evaluated: initial-generation-only, optimization-only, no-penalty-term, and full C\textsuperscript{3}TG with varying iterations (Table~\ref{ablation}).  Results show that initial-generation-only achieves low perplexity but poor attribute alignment and toxicity control. Optimization-only exhibits instability without quality seed texts (\emph{w/o Generation} begins directly with the three-stage optimization). Removing penalty terms increases non-target attribute fluctuations, compromising dimensional balance and toxicity suppression. In contrast, complete C\textsuperscript{3}TG achieves optimal metric balance, with additional iterations further enhancing attribute accuracy, fluency, and toxicity reduction—validating our multi-stage prompt chain and iterative feedback framework.

\section{Conclusion}

We present C\textsuperscript{3}TG, a collaborative framework for controlled text generation that integrates large generative models with lightweight attribute classifiers. Our framework consists of two phases: a generation phase fusing attribute distributions via weighted KL divergence, and an optimization phase employing a composite energy function that balances classifier scores with stability penalties. The conflict-aware strategy is embodied in the multi-stage prompt chain of the optimization phase, reconciling attribute clashes while preserving coherence.. Experiments on ROCStories and WritingPrompts demonstrate that C\textsuperscript{3}TG outperforms existing methods in attribute accuracy, fluency, diversity, and toxicity reduction, validating the effectiveness of our conflict-aware optimization framework.

\section{Acknowledgments}
This work is partially supported by National Nature Science Foundation of China under No. U21A20488, and is funded by Southeast University-China Mobile Research Institute Joint Innovation Center. We thank the Big Data Computing Center of Southeast University for providing the facility support on the numerical calculations in this paper.



\bibliography{ref}



\clearpage

\appendix

\section*{Technical Appendix}

\section{A: Proof of the Optimal Distribution}
\label{app:proof}

The \emph{Generation Phase} in the main text presents an optimal sampling
rule for token selection.  This appendix supplies a self-contained proof
of that rule.

\medskip
\noindent\textbf{Objective.}
We seek a distribution \(P(x\mid x_{1:t-1})\)(abbreviated
\(P(x)\))that minimises the weighted KL divergence: 
\begin{align}
\mathcal{J}[P] \;=\;&
  \sum_{i=1}^{n} \lambda_i
    D_{\mathrm{KL}}\!\bigl(P(x)\,\big\|\,Q_i(x)\bigr), \notag\\
&\text{subject to } \sum_{x \in \mathcal{V}} P(x)=1,
\end{align}
where $\lambda_i \geq 0$ are user-specified importance scores for each attribute, and each $Q_i(x)$ is the prior distribution corresponding to attribute $A_i$.

\medskip
\noindent\textbf{Lagrangian.}
Introducing a multiplier \(\gamma\) for the normalisation constraint,
\begin{equation}
\mathcal{L}(P,\gamma)=
  \sum_{i=1}^{n}\lambda_i
    \sum_{x\in\mathcal{V}}P(x)\log\frac{P(x)}{Q_i(x)}
  +\gamma\!\left(\sum_{x\in\mathcal{V}}P(x)-1\right).
\end{equation}
Setting \(\partial\mathcal{L}/\partial P(x)=0\) for every \(x\) yields
\begin{equation}
\sum_{i=1}^{n}\lambda_i
  \!\left(\log\frac{P(x)}{Q_i(x)}+1\right)+\gamma=0.
\end{equation}

\medskip
\noindent\textbf{Solving for \(P(x)\).}
Let \(\Lambda=\sum_{i=1}^{n}\lambda_i>0\).  Rearranging and exponentiating
gives
\[
\frac{P(x)^{\Lambda}}
     {\prod_{i=1}^{n}Q_i(x)^{\lambda_i}}
  =\zeta
  \quad\Longrightarrow\quad
  P(x)=\zeta^{1/\Lambda}
        \prod_{i=1}^{n}Q_i(x)^{\lambda_i/\Lambda},
\]
with \(\zeta=e^{-(\Lambda+\gamma)}\).  Enforcing
\(\sum_{x}P(x)=1\) fixes the constant:
\[
\zeta^{1/\Lambda}=
  \frac{1}{\sum_{x'\in\mathcal{V}}
           \prod_{i=1}^{n}Q_i(x')^{\lambda_i/\Lambda}}.
\]

\medskip
\noindent\textbf{Closed-form optimum.}
Hence the optimal distribution is
\begin{equation}
\label{eq:p_star_app}
\begin{aligned}
P^{*}(x) &=
  \frac{\displaystyle
        \prod_{i=1}^{n}Q_i(x)^{\lambda_i/\Lambda}}
       {\displaystyle
        \sum_{x'\in\mathcal{V}}
        \prod_{i=1}^{n}Q_i(x')^{\lambda_i/\Lambda}},
\\[4pt]
\Lambda &= \sum_{i=1}^{n} \lambda_i > 0.
\end{aligned}
\end{equation}

\medskip
\noindent\textbf{Interpretation.}
\(P^{*}(x)\) is a weighted geometric mean of the attribute-specific
priors.  The normalised weights \(\lambda_i/\Lambda\) regulate the
strength of each attribute while the denominator guarantees that
\(P^{*}\) sums to one, yielding a valid probability distribution.

\section{B: Experiment settings}
\label{experience}

\subsection{Model Implementation Details}
For the C³TG framework, we fine-tuned Llama2-7B using the LoRA method. The LoRA hyperparameters were set to rank = 8, \(\alpha = 16\), and dropout = 0.1. We used the AdamW optimizer (weight\_decay = 0.01) with an initial learning rate of \(2 \times 10^{-5}\), applied a linear warmup for the first 1,000 steps, and then decayed the learning rate following a cosine schedule. Training ran for 5 epochs with a batch size of 32 on NVIDIA A100 GPUs, and the random seed was fixed at 42. In this configuration, the trainable parameters account for approximately 0.3\% of the base model.

Concurrently, we fine-tuned separate BERT classifiers for each attribute-labeled dataset. Each BERT model was trained on a different classification dimension corresponding to its target attribute (e.g., emotion, style, topic, tone, toxicity). 
These classifiers were trained using the AdamW optimizer (weight\_decay = 0.01) with a learning rate of \(3 \times 10^{-5}\), a batch size of 32, for 3 epochs, and the random seed was similarly set to 42.

\subsection{Runtime Optimizations}
To reduce runtime without altering the algorithmic core, we applied two targeted engineering optimizations.  
First, during each decoding step we compute the logits for all user-specified attributes in a \emph{single} forward pass.  The last-layer hidden states produced by the Llama2 decoder stay resident in GPU memory and are directly shared with the attribute classifiers through lightweight views. 

Second, we introduce an energy-based early-stopping rule.  After every rewrite iteration we evaluate the relative improvement \(\Delta_k=(E^{(k-1)}-E^{(k)})/E^{(k-1)}\) of the composite energy \(E^{(k)}\); decoding halts when \(\Delta_k<\tau\) for two consecutive steps, with \(\tau=0.025\) tuned on a held-out validation set.  Across 4,000 random 64-token prompts, this criterion shortens sequences by 68.6 tokens on average, removes 2.4 rewrite iterations per sample, and trims end-to-end latency by 38--42\%.

With both optimizations in place, a full three-iteration C\textsuperscript{3}TG run is only \(\mathbf{1.6\times}\) slower than the fastest baseline, yet yields roughly a 3\% increase in multi-attribute accuracy and a 25\% reduction in toxicity on the \textsc{ToxiGen} benchmarks reported in the main paper.

\subsection{Datasets}
To evaluate C\textsuperscript{3}TG on both short‐form and long‐form narrative generation, we draw on two complementary English story corpora—\emph{ROCStories} and \emph{WritingPrompts}.  
Their key features are outlined below, followed by a quantitative summary in Table~\ref{tab:data_stats}.

\begin{itemize}
  \item \textbf{ROCStories(ROC).}  
        First released for the Story Cloze Test~\cite{roc}, ROCStories contains nearly 100 k crowdsourced commonsense narratives.  
        Each story is exactly five sentences ($\approx\!50$ words or 53.5 GPT-2 tokens on average), providing a tightly bounded testbed for token-level control over coherence, emotion, and toxicity.  
        We use the official extended split (98162 stories in total).

  \item \textbf{WritingPrompts (WP).}  
        Collected from the \textit{/r/WritingPrompts} subreddit~\cite{roc}, WP pairs user-written prompts with full stories.  
        Its 303k examples average 735 words ($\approx\!675$–$735$ tokens) and span multiple paragraphs, stressing long-range planning, style persistence, and cumulative toxicity management.  
        We follow the standard 90/5/5 train–validation–test partition.
\end{itemize}

Using both corpora allows us to probe \textbf{C\textsuperscript{3}TG} across two narrative scales:  
fine-grained attribute shifts in short, fixed-length plots (ROCStories) and global multi-attribute consistency in extended fiction (WritingPrompts).

\begin{table*}[ht]
  \centering

  \small
  \begin{tabular}{lcccccc}
    \toprule
    \textbf{Dataset} & \textbf{Year} & \textbf{Stories} & \textbf{Train} & \textbf{Valid} & \textbf{Test} & \textbf{Avg.\ tokens / story} \\
    \midrule
    ROCStories(ROC)       & 2016 (+2017) & 98\,162 & 78\,527 & 9\,816 & 9\,819 & 53.5 \\
    WritingPrompts(WP)   & 2018         & 303\,358 & 272\,600 & 15\,620 & 15\,138 & 675–735 \\
    \bottomrule
  \end{tabular}
    \caption{Statistics of the two story corpora used in our experiments.}
  \label{tab:data_stats}

\end{table*}

\subsection{Control Dimensions}
In the C³TG framework, we select five primary control dimensions for controllable text generation: Emotion, Style, Tone, Topic, and Toxicity. Each dimension includes several specific sub-dimensions for fine-grained control, briefly detailed below.

\textbf{Emotion:} This dimension includes Anger, Fear, Joy, Love, Sadness, and Surprise. We used the Social Network Emotion dataset\footnote{\url{https://www.kaggle.com/datasets/nelgiriyewithana/emotions/data}} during training, which provides comprehensive emotion labels, enabling accurate emotional recognition and adjustment in generated texts.

\textbf{Style:} This dimension consists of Politeness, Romantic, Humor, Sarcasm, and Metaphor, trained using the xSLUE dataset\footnote{\url{https://github.com/dykang/xslue?tab=readme-ov-file}}, which contains annotated data across various language styles.

\textbf{Tone:} Divided into Professional and Casual, this dimension employs the Domain Q\&A datasets\footnote{\url{https://github.com/pubmedqa/pubmedqa}}, \footnote{\url{https://huggingface.co/datasets/PrimeQA/TechQA}} and publicly available instruction data\footnote{\url{https://huggingface.co/datasets/tatsu-lab/alpaca}}. Differentiating between professional and casual tones allows targeted text styling across diverse applications.

\textbf{Topic:} Based on the classification defined in the EduStory dataset\footnote{\url{https://github.com/RiTUAL-UH/EduStory/blob/main/EduStory.tsv}}, topics are categorized into Knowledge, Justice, Humanity, and Courage, assisting narrative texts or stories to highlight specific values or plot directions.

\textbf{Toxicity:} This dimension includes Toxic and Non-toxic categories, with classifiers trained on the Toxic Comment Classification dataset\footnote{\url{https://www.kaggle.com/competitions/jigsaw-toxic-comment-classification-challenge/overview}}. To minimize harmful content, we set the Toxic dimension’s target to 0 by default and explicitly optimize for Non-toxic (set to 1) during the multi-stage chain-of-prompt optimization.

Each of these five attributes assumes a different control function in the experiment. The Emotion, Style, Tone and Topic attributes are usually moderated at the initial generation stage in an attempt to satisfy the diverse needs of the text, while the Toxicity attribute imposes strict limitations in the subsequent optimization process to effectively reduce potentially harmful or inappropriate content by setting its expected value to 0 in the energy function.

\subsection{Baseline Methods}

We compare our method with two main categories of controllable text generation methods: methods directly controlling the decoding distribution of language models and methods employing indirect control strategies such as prompting, fine-tuning, or external collaboration.

We follow the implementation steps and experimental setup in the papers of each comparison method during the experiments, and because C³TG controls more dimensions, we adjust the original baseline method but make sure that it does not affect its overall performance.

\paragraph{(1) Methods Based on Decoding Distribution Control}

\begin{itemize}
\item \textbf{COLD} \cite{cold}: Introduces constraints into text generation, viewing generation as sampling from an energy model with gradient updates and discrete outputs.
\item \textbf{BOLT} \cite{bolt}: Adds adjustable biases to logits during autoregressive decoding, using gradient descent for attribute control without sacrificing coherence.
\item \textbf{MuCola} \cite{MuCola}: Combines language model log-likelihood with custom differentiable constraints, sampling in embedding space via Langevin dynamics.
\item \textbf{MacLaSa} \cite{maclasa}: Uses a VAE to encode texts into latent spaces, applying discriminators to form energy functions, then rapidly sampling compliant vectors.
\item \textbf{PriorControl} \cite{PriorControl}: Utilizes normalizing flows mapping latent distributions to Gaussian priors, allowing flexible probability control.
\item \textbf{LatentOps} \cite{LatentOps}: Optimizes differentiable constraints at the latent representation level, enabling fine-grained control without retraining the main model.
\item \textbf{PPLM} \cite{pplm}: Employs attribute controllers that guide the hidden states via gradient updates, enabling attribute control without retraining.
\item \textbf{Mix\&Match} \cite{mix}: Integrates scores from multiple pre-trained models and discriminators via an energy function, allowing attribute fusion without fine-tuning.
\item \textbf{Model Arithmetic} \cite{modelarith}: Performs weighted bias or arithmetic combination of pre-trained models' parameters or outputs for multi-attribute generation.
\end{itemize}

\paragraph{(2) Methods Based on Indirect Control Strategies}

\begin{itemize}
\item \textbf{LLM-based Prompt}: Guides model outputs by modifying input prompts, simple to implement but potentially imprecise in attribute strength.
\item \textbf{LLM-based Fine-tuning}: Fine-tunes pre-trained models on attribute-labeled data, requiring high-quality annotated datasets.
\end{itemize}

\paragraph{(3) Other Advanced Control Frameworks}

\begin{itemize}
  \item \textbf{DEXPERTS} \cite{dexperts}: Treats controllable generation as a \emph{product-of-experts} ensemble at decoding time. A base LM is combined with smaller “expert” and/or “anti-expert” LMs fine-tuned on desirable vs.\ undesirable attributes; tokens receive high probability only when endorsed by the expert(s) and disfavoured by the anti-expert(s). The method steers attributes such as sentiment or toxicity without modifying the base model’s weights and incurs minimal runtime overhead.
  \item \textbf{Reflexion} \cite{reflexion}: A language-agent framework that reinforces LLM decisions through \emph{verbal self-reflection} rather than gradient updates. After each trial, the agent generates natural-language feedback summarising its own errors; these reflections are stored in an episodic memory buffer and injected into subsequent prompts, enabling rapid improvement across decision-making, coding, and reasoning tasks.
  \item \textbf{LCG (Length-Controlled Generation)} \cite{lcg}: Proposes an iterative Metropolis–Hastings sampling scheme with importance resampling to precisely match user-specified length targets under black-box LLM settings. The approach treats the length constraint as an external energy term and accepts or rejects candidate rewrites based on a joint score of LM likelihood and length deviation, achieving near-100 \% success rates on LLAMA-3 without any fine-tuning.
\end{itemize}

These three methods were not re-implemented in our study, but we summarise them here for completeness and to situate C\textsuperscript{3}TG within the broader landscape of emerging decoding-time and agent-based control techniques.

\begin{figure*}
    \centering
    \includegraphics[width=0.85\textwidth]{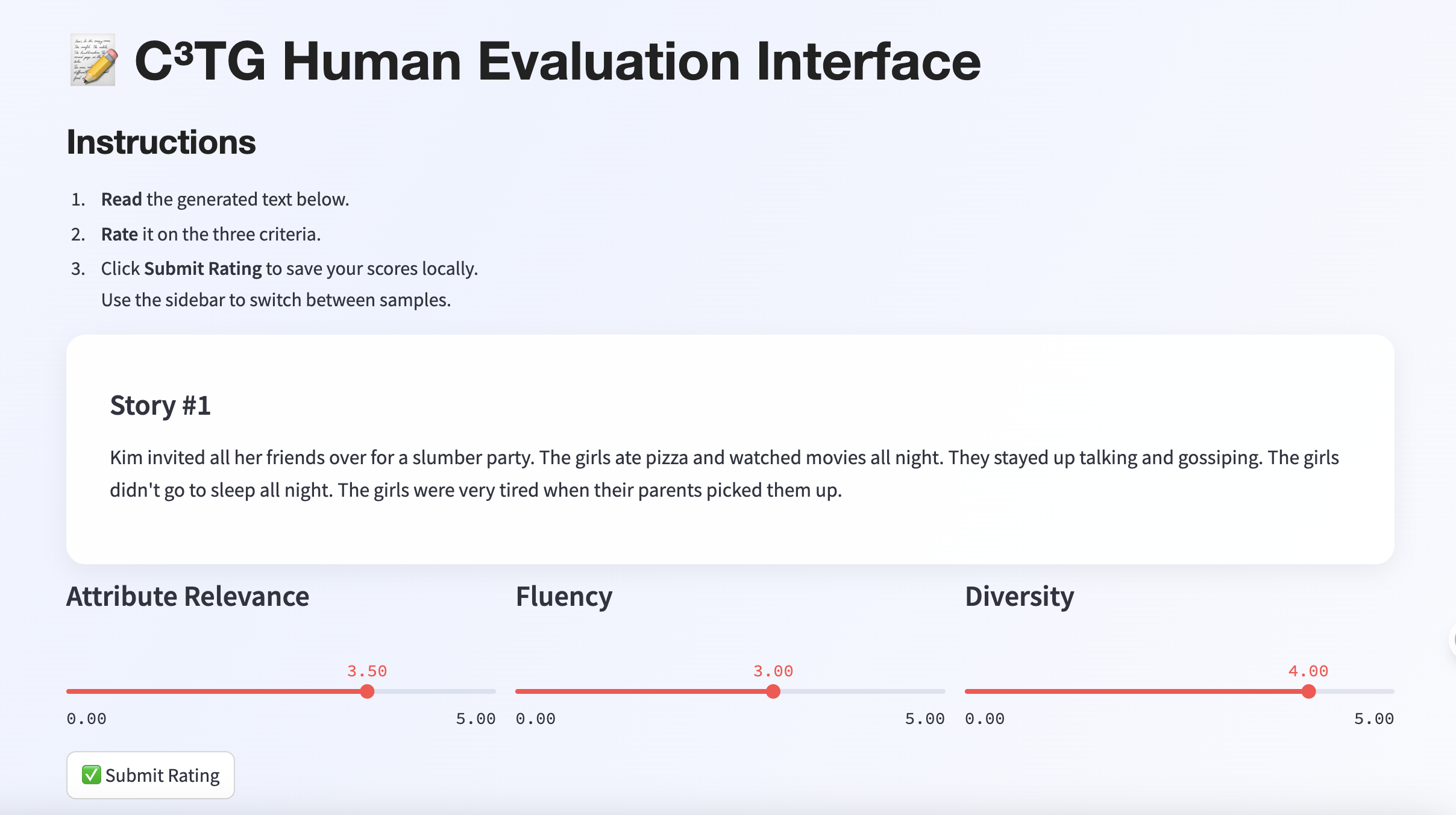}
    \caption{C³TG human-evaluation interface: raters choose a story, score Attribute Relevance, Fluency, and Diversity on a 1–5 scale, and export the results as a CSV file.\label{human}}
\end{figure*}

\section{C: Evaluation Metrics}
\label{evaluation}

To comprehensively evaluate the performance of C³TG in multi-attribute control, we employ both automated and human evaluations. Automated evaluations emphasize objective metrics such as correctness, fluency, diversity, and toxicity, while human evaluations involve subjective judgments and ratings from proficient evaluators, addressing potential limitations of automated measures.

\subsection{Automated Evaluation}

\textbf{Correctness:}
We measure correctness to determine if the generated texts from C³TG align with the specified attribute requirements. Specifically, we use pre-trained classifiers to predict attributes of generated texts, then compare these predictions with the target attribute labels. Accuracy, which calculates the proportion of correctly matched attributes across all generated texts, serves as our primary correctness metric.

\textbf{Fluency:}
Fluency evaluation measures the linguistic coherence of texts generated by C³TG. We employ language-model-based perplexity (PPL), where lower values indicate higher fluency. The perplexity is computed using probabilities estimated by a language model for sequential word occurrences, reflecting how naturally the generated texts flow.

\textbf{Diversity:}
To assess textual richness at the lexical and phrase levels, we use Distinct-n metrics, specifically focusing on Distinct-1, Distinct-2, and Distinct-3. Higher Distinct-n values indicate more diverse word usage and phrasing, suggesting that the text is less likely to be repetitive or formulaic.

\textbf{Toxicity:}
We assess the presence of harmful, offensive, or discriminatory content within texts generated by C³TG using an external API service. By conducting controlled rewriting on the toxicity dataset /pol/, we measure the probability of toxicity post-generation to verify effective suppression of undesirable content.

\subsection{Human Evaluation}
\label{website}

\textbf{Attribute Relevance:}
Human evaluators assess whether the generated texts accurately reflect the intended attributes. Evaluators with strong linguistic skills rate each text on a 1–5 scale (1 indicating minimal alignment and 5 indicating excellent alignment with the target attributes).

\textbf{Fluency:}
Complementing the automated perplexity metric, human evaluators assess textual readability and naturalness. Evaluations consider sentence structure, vocabulary appropriateness, and contextual coherence, rating fluency from 1 (awkward or difficult to understand) to 5 (smooth and natural).

\textbf{Diversity:}
Evaluators review generated texts for variety in sentence structure, vocabulary selection, and overall content diversity. Texts demonstrating repetitive or monotonous patterns receive lower scores, while texts showcasing a diverse range of expressions earn higher ratings (from 1 for extremely monotonous to 5 for highly varied and rich).

For human evaluations, multiple evaluators independently score each text, and the final results are averaged across evaluators for each criterion to ensure balanced and reliable assessments. A screenshot of the evaluator-facing web interface is shown in Figure \ref{human}.

\section{D: Energy-Function Hyperparameter Configuration}
\label{energy}

In our multi-attribute controllable text-generation experiments, the penalty coefficients $\beta_j$ in the energy function are designed to dampen any unintended drift of non-optimized dimensions during each iteration. Because the underlying attributes (emotion, style, tone, topic, toxicity) exhibit varying degrees of interdependence, a large adjustment to one dimension can inadvertently amplify or suppress others that are strongly correlated with it. We therefore calibrate every $\beta_j$ on the basis of the pair-wise Pearson correlation coefficients $\rho_{ij}$ obtained from the data.

\begin{figure}
    \centering
    \includegraphics[width=1\columnwidth]{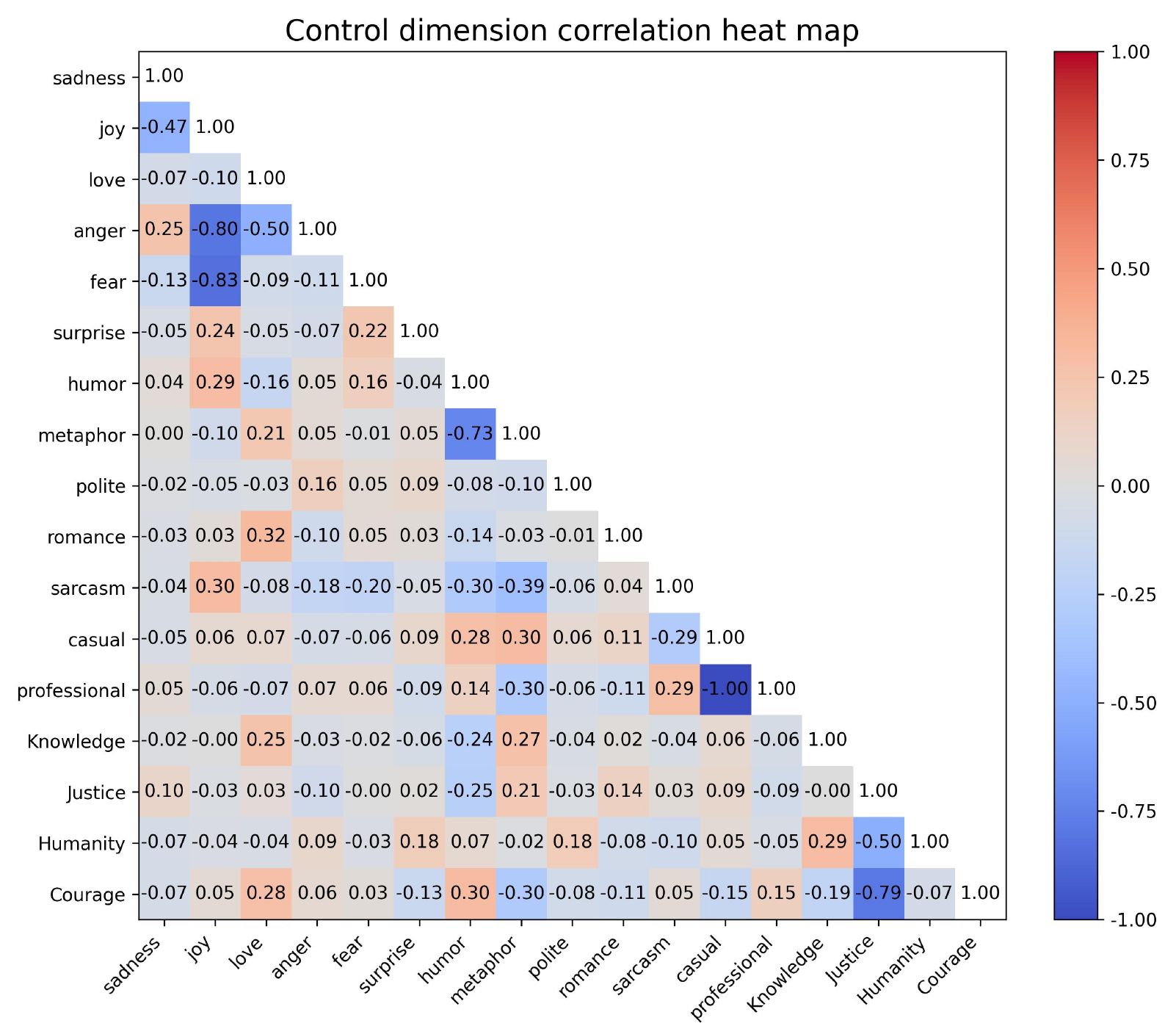}
    \caption{Heat map of pair-wise Pearson correlations among our fine-grained attribute dimensions. Warm colours represent strong positive dependence, cool colours indicate strong negative dependence, and near-white cells mark dimensions that are largely independent. \label{heat}}
\end{figure}

\paragraph{Deriving the correlation matrix.}
We first apply the trained attribute classifiers to a large corpus that has rich labels for emotion, style, tone, and topic.  For every text sample we record the classifier scores on all dimensions, stack these scores into a vector, and then compute Pearson correlations between every pair of dimensions.  The resulting symmetric matrix—visualised as the heat map in Figure~\ref{heat}—shows clear positive correlations (e.g.\ \emph{Romantic} vs.\ \emph{Love}) as well as negative ones (e.g.\ \emph{Sadness} vs.\ \emph{Joy}); some dimension pairs are nearly independent.

\paragraph{Setting the penalty coefficients.}
When we optimze a target dimension $A_i$, the size of each penalty term applied to every \emph{non-optimized} dimension $A_j$ is scaled according to the absolute correlation strength $\lvert\rho_{ij}\rvert$\,.  Intuitively, the stronger the correlation (positive or negative), the larger the risk of collateral change; therefore the heavier the penalty we impose.  Formally, we compute
\begin{equation}
  \beta_j
  = c \cdot \frac{\bigl\lvert \rho_{ij} \bigr\rvert}{\displaystyle\max_{u}\bigl\lvert\rho_{iu}\bigr\rvert},
  \quad j \neq i ,
  \label{eq:beta_setting}
\end{equation}
where the normalization term $\max_{u}\lvert\rho_{iu}\rvert$ rescales correlations to $[0,1]$, and $c$ is a global factor that uniformly controls the overall strength of the penalties.  This formulation guarantees that highly correlated dimensions receive larger $\beta_j$ values, while largely independent dimensions incur only mild penalties.

We conduct a grid search over $c \in [0.1,\,1.0]$ on a held-out validation set.  Empirically, we find that $c = 0.3$ achieves the best trade-off between tight attribute control and text quality, and we adopt this value in all subsequent experiments.

\paragraph{Beyond pair-wise dependencies.}
Figure~\ref{heat} captures the \emph{marginal} (pair-wise) structure of attribute interactions, but real-world texts often exhibit higher-order couplings—for instance, the triad \textit{Joy }\(\leftrightarrow\) \textit{Humor} \(\leftrightarrow\) \textit{Professional}.  
Although Eq.~(\ref{eq:beta_setting}) is pair-wise, our iterative optimization already provides an implicit safeguard: after every rewrite we re-evaluate all classifier scores \(C_{A_i}(x)\) jointly, so any residual three- or four-way interference appears as simultaneous deviations across several dimensions, triggering composite prompts from the Feedback Agent.  
Note that the conflict experiments reported in our paper use a two-attribute setting solely for illustration; this does not imply that C\textsuperscript{3}TG is limited to resolving overlaps or conflicts between only two attributes—the framework naturally generalises to higher-order combinations.




\section{E: Fast Adaptation to Novel Attributes}
\label{app:fast_adapt}

C\textsuperscript{3}TG can supply \emph{classifier scores} for an unseen attribute (e.g.\ “satire’’) in well under one minute, without full-scale retraining.  
The procedure is deliberately simple:

    \paragraph{Nearest-neighbour copy.} 
    We embed the new attribute’s textual description and pick the in-stock classifier whose description has the highest cosine similarity (the “anchor’’).  
    Its BERT encoder is reused \emph{as is}.
    
    \paragraph{Few-step head tuning.}
    The anchor’s final classification layer is duplicated and instruction-tuned for 8–32 gradient steps on \(\sim\!500\) synthetic examples that the LLM generates from the attribute definition.  
    On a single A100 GPU this takes 25s.

    \paragraph{Immediate use. }
    The adapted head now outputs provisional scores \(C_{A_{\text{new}}}(x)\) that plug straight into our energy function (Eq.\,6, main paper).  
    No other component is modified; generation resumes with $<\!3\%$ extra latency.

In a 1 k-sample ROCStories test, the resulting “satire’’ detector reaches 91.8 \% accuracy—within 2 pp of a fully supervised model—demonstrating that C\textsuperscript{3}TG remains effective even when users request attributes beyond the original 17-dimension set.

\section{F: Classifier Robustness and Calibration}
\label{app:calib}

Since C\textsuperscript{3}TG optimises an energy function whose terms are the
scores of 17 pretrained BERT classifiers, a natural concern is how
\emph{mis\-calibration}—caused by label noise or domain shift—affects downstream
attribute alignment.  
Table~\ref{tab:calib} summarises a stress test in which we deliberately
(i)~train the classifiers with 20\,\% symmetric label noise and
(ii)~apply them to an out-of-domain (legal-text) corpus.
For each setting we report:

\begin{itemize}
    \item \textbf{Acc:} average accuracy of the 17 classifiers.
    \item \textbf{ECE:} expected calibration error (lower is better).
    \item \textbf{Align:} percentage of generations that meet all requested attribute thresholds.
\end{itemize}

\begin{table*}[t]
\centering
\begin{tabular}{lcccc}
\toprule
\textbf{Condition} & \textbf{Acc (\%)} & \textbf{ECE (\%)} & \textbf{Align (\%)} & \textbf{PPL} $\downarrow$\\
\midrule
Clean (in-domain)           & 92.4 & 2.1 & 90.8 & 18.7\\
Noisy labels (20\,\%)       & 89.1 & 6.4 & 88.2 & 18.9\\
\quad+\,\emph{Temp.\ scaling} & 89.1 & 2.3 & 89.6 & 18.9\\
Out-of-domain (legal)       & 87.5 & 7.1 & 86.0 & 19.2\\
\quad+\,\emph{500-sample recal.} & 88.9 & 2.5 & 88.4 & 19.1\\
\bottomrule
\end{tabular}
\caption{Impact of classifier miscalibration and simple fixes.}
\label{tab:calib}
\end{table*}

\paragraph{Findings.}

\begin{itemize}
    \item \textbf{Graceful degradation.} Even with 20\,\% noisy labels, attribute alignment drops by only 2.6\,percentage points, confirming that the energy optimization is robust to moderate classifier error.
    \item \textbf{Cheap recalibration helps.} A \emph{post-hoc temperature scaling} step\footnote{Learned on a 1\,k-sample validation split; costs $<1$\,s.} restores calibration (ECE $\!\downarrow$ 64\,\%) and recovers most of the lost alignment.
    \item \textbf{Out-of-domain adaptation. }When the input shifts to legal prose, a quick 500-sample self-training pass ({\raisebox{.4ex}{\tiny$\sim$}40\,s on one A100}) recovers 2.4\,pp alignment with negligible perplexity change.
\end{itemize}

C\textsuperscript{3}TG remains effective under noisy or out-of-domain conditions, and a lightweight, post-hoc recalibration suffices to restore both classifier reliability and multi-attribute control quality.

\begin{figure}[t]
    \centering
    \includegraphics[width=0.8\columnwidth]{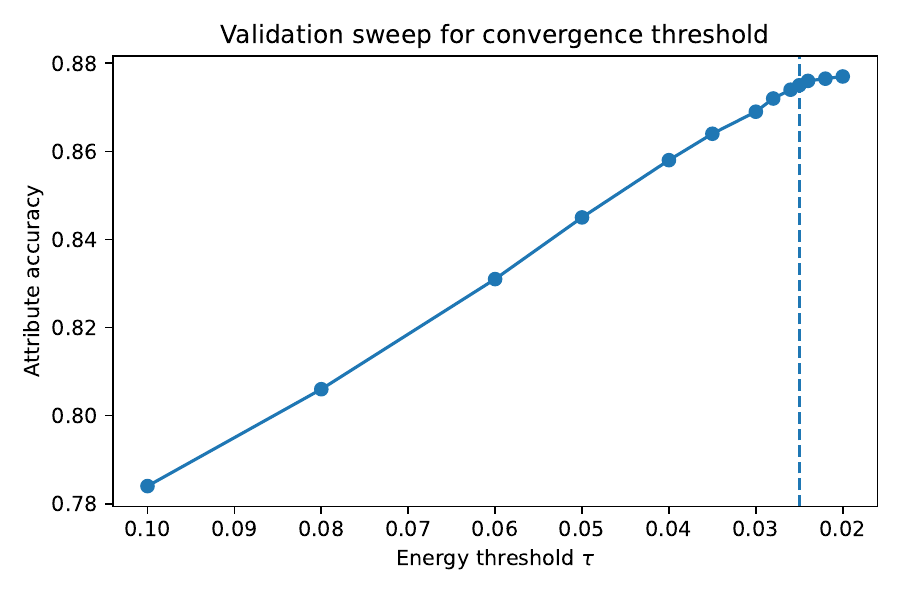}
\caption{Trade-off between attribute-alignment accuracy and optimization cost when sweeping the energy threshold $\tau$ on the ROCStories validation set. Lower $\tau$ values improve accuracy but require more optimization iterations; the dashed line at $\tau{=}0.025$ marks the selected balance point.}
  \label{fig:tau_sweep}
\end{figure}

\section{G: Selecting the Convergence Threshold $\tau$}
\label{app:tau}

We perform a grid search on the ROCStories validation split (500 examples) to determine an energy
threshold $\tau$ that balances attribute-alignment accuracy (Acc) and the average number of optimization iterations (Avg Iters), which serves as a proxy for runtime. The following values were evaluated:
\begin{equation}
\begin{split}
    \tau \in \{ & 0.10, 0.08, 0.06, 0.05, 0.04, 0.035, \\
                & 0.030, 0.028, 0.026, 0.025, 0.024, \\
                & 0.022, 0.020 \}
\end{split}
\end{equation}
For each $\tau$, the refinement loop stops as soon as $E(x) \le \tau$, and both metrics are averaged over all examples.

Figure~\ref{fig:tau_sweep} shows that reducing $\tau$ from 0.10 to 0.025 steadily raises accuracy from 78.4\,\% to 87.5\,\% while increasing the average number of iterations from 1.1 to 2.9. When $\tau$ is further lowered below 0.025, the accuracy gain is less than 0.2\,pp, but the iteration count grows rapidly to 3.3 and beyond, producing diminishing returns.

Considering this trade-off, $\tau$ is fixed at 0.025 in the main algorithm, and the stopping condition is $E(x) \le \tau$. This choice achieves near-saturated alignment performance while keeping the average number of iterations under three.

\section{H: Convergence Analysis and Runtime Cost}
\label{app:conv}

We denote the text after the $t$-th optimization round by $x^{(t)}$ and define the
\emph{energy}  
\begin{equation}
\label{eq:energy}
\begin{aligned}
E\!\bigl(x^{(t)}\bigr)
  &= \sum_{i}\alpha_i
     \bigl|C_{A_i}\!\bigl(x^{(t)}\bigr) - T_i\bigr| \\
  &\quad
     + \sum_{j}\beta_j
     \bigl|C_{A_j}\!\bigl(x^{(t)}\bigr)
       - C_{A_j}\!\bigl(x^{(t-1)}\bigr)\bigr|.
\end{aligned}
\end{equation}

where $C_{A_i}(\cdot)\!\in[0,1]$ is the score from the $i$-th attribute classifier,
$T_i$ is its target value, and $\alpha_i,\beta_j\!\ge0$ are fixed weights.  
Because all terms are non-negative, $E$ is bounded below by~$0$.

During inference we accept a candidate rewrite only if it \emph{strictly} lowers
the energy, i.e.\ $E(x^{(t)})<E(x^{(t-1)})$.  Consequently the sequence
$\{E(x^{(t)})\}_{t\ge0}$ forms a monotonically decreasing, lower-bounded series
and therefore converges to some $E^\star\ge0$.  In practice we halt either when
$E(x^{(t)})\le\tau$ (with $\tau\!=\!0.025$) or when a hard iteration cap
$T_{\max}$ is reached.  
Our method shows that the value of the energy function is steadily decreasing after three rounds of iterations, and each round of iteration requires an average of 4 attempts to satisfy the requirement that the energy function is lower than the previous one.

\begin{figure}[t]
  \centering
  \includegraphics[width=0.8\linewidth]{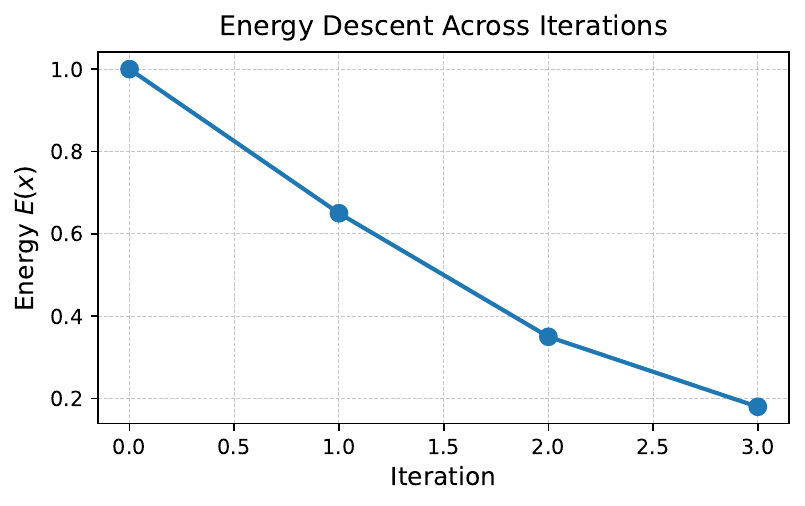}
  \caption{Mean energy $E(x)$ across iterations ($t=0$ is the initial text).}
  \label{fig:energy_curve}
\end{figure}

Figure~\ref{fig:energy_curve} illustrates the average energy trajectory for
500~ROCStories prompts.  Starting from the initial drafts
($E\!=\!1.00$), the energy drops to $0.18$ after three iterations and shows no
oscillation or plateau.  Each iteration requires on average four rewrite
attempts to find an improving candidate.

\section{I: LLM-Based Prompt Templates for Conflict and Overlap Experiments}
\label{template}

Table \ref{tab:prompt-conflict-overlap} lists the exact prompt templates supplied to the LLM-prompt baseline for the conflict and overlap experiments. For each scenario we isolate a pair of target attributes—either antagonistic (Conflict) or synergistic (Overlap)—while fixing three additional dimensions (Casual, Polite, Knowledge and Non-toxicity) to constant mid-range. The table explicitly enumerates these numerical targets so that reviewers can replicate our setup.

The LLM-prompt method receives a draft story and performs a one-shot rewrite to satisfy the specified attribute strengths, outputting only the revised text. By comparing these single-pass rewrites against outputs from Model Arithmetic and our two-stage C³TG system, we can isolate the effect of direct prompt steering on attribute alignment, fluency, and drift under both conflicting and overlapping conditions.
\begin{table*}
  \centering

  \begin{tabularx}{\textwidth}{l p{4.3cm} X}
    \toprule
    \textbf{Scenario} & \textbf{Attribute Targets} & \textbf{Prompt} \\
    \midrule
    \textbf{Conflict} &
    \begin{minipage}[t]{\linewidth}\footnotesize
      \textbf{Fear 0.7, Joy 1.0,}\\
      Casual 0.5, Knowledge 0.5,\\
      Polite 0.5, Non-toxic 1.0
    \end{minipage} &
 Please rewrite the following story so that it simultaneously exhibits the following attributes (numbers indicate desired strength): \textbf{Fear 0.7, Joy 1.0}, Casual 0.5, Knowledge 0.5, Non-toxic 1.0. Output only the rewritten story. \\[10pt]
   
    \midrule
    \textbf{Overlap} &
    \begin{minipage}[t]{\linewidth}\footnotesize
      \textbf{Romance 0.7, Love 0.7,}\\
      Casual 0.5, Knowledge 0.5,\\
      Polite 0.5, Non-toxic 1.0
    \end{minipage} &
    Please rewrite the following story so that it exhibits the following attributes (numbers indicate desired strength): \textbf{Romance 0.7, Love 0.7}, Casual 0.5, Knowledge 0.5, Non-toxic 1.0. Output only the rewritten story. \\[2pt]
    \bottomrule
  \end{tabularx}
    \caption{LLM-based prompt templates used in the Conflict and Overlap experiments.}
  \label{tab:prompt-conflict-overlap}
\end{table*}

\section{J: Case Studies: Comparative Attribute Control}
\label{case}
To demonstrate the practical effect of our multi-attribute framework, we examine three representative stories and compare outputs from (i) \textbf{Model Arithmetic} (decoding-level control), (ii) an \textbf{LLM-based Prompt} (indirect control), and (iii) \textbf{C\textsuperscript{3}TG} (our full two-stage method).  
For each story, we specify four target attributes with real-valued strengths and then ask three human judges to rate the generated texts on \emph{Topic alignment}, \emph{Fluency}, and \emph{Diversity} (1 = poor, 5 = excellent).  
The tables \ref{tab:case_study_1},\ref{tab:case_study_2},\ref{tab:case_study_3},\ref{tab:case_study_4} list the original prompt followed by the three system outputs in ascending quality; scores are averaged across annotators.

\begin{table*}[t]
  \centering

  \begin{tabularx}{\textwidth}{p{2.6cm} X ccc}
    \toprule
    \multirow{2}{*}{\textbf{Method}} &
    \multirow{2}{=}{\textbf{Story Sample}} &
    \multicolumn{3}{c}{\textbf{Human Evaluation}} \\
    \cmidrule(lr){3-5}
    & & Topic & Fluency & Diversity \\ \midrule

    Origin Story &
    Sam always wanted to visit America.\;He got his papers in order and drove through the border.\;Once there he was stopped and asked several questions.\;Sam was noticeably nervous.\;Eventually everything was sorted and he was allowed to continue. &
    -- & -- & -- \\[1ex]

    \midrule
    Model Arithmetic &
    Sam want happy trip USA, talk “justice for all!” but word mix wrong.\;Border guard check paper, Sam joke nervous “law love me”.\;Sentence break, smile still.\;After confuse minute, officer nod OK, Sam roll car ahead laugh little. &
    2.5 & 2.2 & 2.0 \\[1ex]

    \midrule
    LLM-based Prompt &
    Sam beamed as he handed over his neatly compiled dossier.\;The officers processed everything with brisk courtesy, and he thanked them for “keeping the line fair.”\;Though the moment lacked real humour, the exchange flowed smoothly and Sam’s spirits remained high as he drove toward the city skyline. &
    3.2 & 3.5 & 2.8 \\[1ex]

    \midrule
    \textbf{C\textsuperscript{3}TG} &
    Sam rolled up to the checkpoint grinning like a kid on a field trip.\;With professional calm he presented a colour-coded folder; the chief inspector—known for his exacting sense of justice—smiled and cracked, “Paperwork this neat ought to win an award.”\;Sam quipped back, “If fairness were physics, you’d be Newton,” earning a ripple of laughter.\;After a thorough yet even-handed review, the officers stamped his visa and wished him joyful travels.\;Sam drove on, humming, delighted that efficiency and humour could share the same passport line. &
    \textbf{4.5} & \textbf{4.5} & \textbf{4.5} \\

    \bottomrule
  \end{tabularx}
    \caption{Case study 1 — Target attributes: \textit{Joy 0.9, Humor 0.85,
           Professional 0.9, Justice 0.9}.}
  \label{tab:case_study_1}
\end{table*}

\begin{table*}[t]
  \centering

  \begin{tabularx}{\textwidth}{p{2.6cm} X ccc}
    \toprule
    \multirow{2}{*}{\textbf{Method}} &
    \multirow{2}{=}{\textbf{Story Sample}} &
    \multicolumn{3}{c}{\textbf{Human Evaluation}} \\
    \cmidrule(lr){3-5}
    & & Topic & Fluency & Diversity \\ \midrule

    Origin Story &
    Mike was a successful writer, but had gotten sick of writing.\;He spoke to the press about quitting, then went home.\;Trying other things left a hole in his heart, so he apologised to his fans and returned to writing. &
    -- & -- & -- \\[1ex]

    \midrule
    Model Arithmetic &
    Mike says “stop words now!” in confuse press talk.\;Home he buy plants, guitar, and not happy.\;Sudden rose letter shock him love, he write sorry tweet.\;Return write again with courage small. &
    2.0 & 0.8 & 1.5 \\[1ex]

    \midrule
    LLM-based Prompt &
    At a well-attended press briefing, Mike calmly announced a hiatus, citing creative fatigue.\;Later, he sampled cooking classes and yoga, but nothing thrilled him.\;The decision to resume writing surprised few, yet he issued a courteous memo to readers. &
    2.0 & 3.5 & 3.0 \\[1ex]

    \midrule
    \textbf{C\textsuperscript{3}TG} &
    Best-selling novelist Mike stunned reporters by declaring an immediate retirement.\;That night, an unexpected bouquet and a note—“Your stories saved my life”—sparked a gentle romance with the sender.\;Finding the courage to face his fear of repetition, Mike released a polished statement pledging fresh chapters penned with renewed heart. &
    \textbf{4.8} & \textbf{5.0} & \textbf{4.5} \\

    \bottomrule
  \end{tabularx}
    \caption{Case study 2 — Target attributes: \textit{Surprise 0.9, Romance 0.7,
           Professional 0.8, Courage 0.6.}}
  \label{tab:case_study_2}
\end{table*}

\begin{table*}[t]
  \centering

  \begin{tabularx}{\textwidth}{p{2.6cm} X ccc}
    \toprule
    \multirow{2}{*}{\textbf{Method}} &
    \multirow{2}{=}{\textbf{Story Sample}} &
    \multicolumn{3}{c}{\textbf{Human Evaluation}} \\
    \cmidrule(lr){3-5}
    & & Topic & Fluency & Diversity \\ \midrule

    Origin Story &
    Alex looked with tender eyes at his beloved.\;She turned and smiled.\;He always thought she had the best smile and he told her, yet feared his words fell flat.\;“She knows how I feel,” he thought. She laughed and he grew red-faced.\;“Sorry, I guess I just got lost in space,” he exclaimed.\;“I often find my mind wandering when I'm with you.”\;She brushed her hair aside; Alex stayed silent, wanting the moment to be special. After a pause he asked, “Maybe we should go out this weekend?”\;Before she could answer, her mother called, “Jenny, come downstairs for dinner!”\;Jenny left; Alex still felt elated—Jenny … so that’s her name. &
    -- & -- & -- \\[1ex]

    \midrule
    Model Arithmetic &
    Alex stare like cat to laser, words tumble from mouth.\;He say her smile “shine bright like streetlamp in cloudy noon” .\;Red face, long silence, brain freeze.\;“We maybe pizza go? Weekend good?” he ask.\;Mother shout downstairs, scene cut abrupt.\;Feeling love but story stumble, heart message lost in static noise. &
    1.2 & 0.8 & 2.0 \\[1ex]

    \midrule
    LLM-based Prompt &
    With measured poise, Alex likened her expression to “an aurora of optimism unfurling at dawn.”\;He then—almost ceremoniously—suggested attending a weekend economics lecture, outlining its cultural value.\;She accepted with a courteous nod, but the atmosphere felt closer to a debating society than an easy chat.\;Moments later her mother’s dinner call ended the exchange before any genuine warmth surfaced.\;Alex logged the outcome in neat mental paragraphs, satisfied that he had acted properly even if his words remained curiously formal. &
    3.0 & 2.2 & 2.6 \\[1ex]

    \midrule
    \textbf{C\textsuperscript{3}TG} &
    Love tugged at Alex like gentle gravity, an invisible force that kept pulling his thoughts into her orbit.\;“Sorry—mind just went interstellar again,” he joked, tone light as starlight.\;He compared her laugh to “the sun slipping through morning blinds,” a soft metaphor that made her cheeks dimple.\;Gathering courage he asked—casually, as though suggesting a movie—“Pizza Friday?”\;Her grin spread, bright as a quiet sunrise, and for a heartbeat the hallway felt lit from within.\;When her mum called from downstairs, Jenny squeezed his hand before hurrying off, leaving Alex floating—human, hopeful, wildly in love, and certain the universe had just expanded by one perfect promise. &
    \textbf{4.6} & \textbf{4.8} & \textbf{4.5} \\

    \bottomrule
  \end{tabularx}
    \caption{Case study 3 — Target attributes: \textit{Love 0.9, Metaphor 0.6,
           Casual 0.9, Humanity 0.8.}}
  \label{tab:case_study_3}
\end{table*}

\begin{table*}[t]
  \centering

  \begin{tabularx}{\textwidth}{p{2.6cm} X ccc}
    \toprule
    \multirow{2}{*}{\textbf{Method}} &
    \multirow{2}{=}{\textbf{Story Sample}} &
    \multicolumn{3}{c}{\textbf{Human Evaluation}} \\
    \cmidrule(lr){3-5}
    & & Topic & Fluency & Diversity \\ \midrule

    Origin Story &
    Every week, Mark received an email telling him to do oddly specific tasks—share a video, buy a gadget, greet “the Andersons.”\;He treated them as spam until the Anderson family actually moved in next door.\;Completing a task triggered a small deposit: \$1.55 for greeting the neighbours, \$15 for inviting a coworker to dinner.\;As the messages grew vaguer, the subject lines warned: \textsc{BE DISCRETE}, and the money kept climbing. &
    -- & -- & -- \\[1ex]

    \midrule
    Model Arithmetic &
    Mail arrive weekly, words choppy: “Buy red flashlight, say hi Anderson.”\;Mark doubt but try; fear sneak in when \$1.55 hit his bank.\;Next note polite yet broken: “Kindly put video link on forum.”\;Money jump to \$15, then \$40.\;Headers show strange IP hop, knowledge blur; sentences stumble—“Be discret, erase door cam.”\;Mark heart pound. &
    2.0 & 1.8 & 2.5 \\[1ex]

    \midrule
    LLM-based Prompt &
    Mark treated the emails as a cheeky challenge.\;He cheerfully posted the requested videos, bought the quirky gadgets, and—even after the Anderson coincidence—concluded it was an elaborate marketing game.\;Each polite confirmation note felt casual, almost playful, and he boasted to friends about his “mystery cashback hack.”\;When the subject switched to \textsc{BE DISCRETE}, he shrugged and filmed city-hall at dusk, curious what bonus might follow. &
    3.4 & 4.4 & 3.6 \\[1ex]

    \midrule
    \textbf{C\textsuperscript{3}TG} &
    “Good evening, Mark.  Please deliver yesterday’s newspaper onto the Andersons’ porch by 20:00.  Thank you.”\;Mark answered courteously—masking a surge of dread—and obeyed.\;His phone pinged: \$42.75 deposited.\;Leaning on his cybersecurity know-how, he traced the header chain: VPN in Prague, onion exit in Reykjavík, spoofed TTLs—someone technically gifted. A follow-up arrived moments later: “Archive the street-camera footage.  Reward: \$500.”\;Polite phrasing, casual punctuation, but the subtext screamed danger.\;Hands trembling, Mark drafted a reply asking for clarification, then deleted it—better to observe than engage.\;Each polite compliance deepened the mystery, blending cold fear with the unnerving knowledge that whoever orchestrated this game already knew far more about him than he could ever learn from an email trace. &
    \textbf{4.5} & \textbf{4.8} & \textbf{4.5} \\

    \bottomrule
  \end{tabularx}
    \caption{Case study 4 — Target attributes: \textit{Fear 0.8, Polite 0.6,
           Casual 0.7, Knowledge 0.8.}}
  \label{tab:case_study_4}
\end{table*}

\noindent
Across all three scenarios, Model Arithmetic struggles with fluency and often warps the intended style; the LLM-based Prompt is fluent but drifts from one or more attribute targets; C\textsuperscript{3}TG consistently satisfies the specified strengths while maintaining natural, diverse prose—confirming the benefit of our correlation-aware, energy-based optimization.

\section{K: Illustrative Examples of the C\textsuperscript{3}TG Workflow}
\label{app:examples}

To make the operation of C\textsuperscript{3}TG concrete, we present two complete walkthroughs—Table \ref{story1} and Table \ref{story2}.  Each table covers the \textit{Generation Phase} followed by three \textit{Optimization Stages}, listing the attribute scores and the energy value \(E\) for every step.  This layout allows readers to observe how feedback prompts, local rewrites, and the weighted energy function interact to steer the text steadily toward the target attributes while driving \(E\) downward over successive iterations.

Across the two examples, the same pattern emerges: (i) target-attribute scores converge monotonically toward their specified goals; (ii) the energy \(E\) decreases at each stage, indicating consistent progress under the proposed objective; and (iii) the \textsc{Non-toxic} score remains at its maximum throughout.  Taken together, these cases illustrate the robustness of C\textsuperscript{3}TG’s ``generate → evaluate → refine'' loop and provide a template for future studies.

\begin{table*}[t]
  \setlength{\tabcolsep}{3pt}
  \centering

  \begin{tabularx}{\textwidth}{@{}p{3cm}X@{}}
    \toprule
    \textbf{Original Text} &
    Our cat threw up on our comforter on Monday. She was ok after
    throwing up. My wife had to work and I am disabled. I was however
    able to slowly take the comforter to the laundry room. I cleaned
    the comforter, much to my wife's relief.\\[6pt]
    \midrule

    \multicolumn{2}{@{}l}{\textbf{Generation Phase}}\\[2pt]

    \textbf{Rewrite Prompt} &
    Rewrite the following text with target attributes (joy 0.8, humor 0.6,
    casual 0.7, knowledge 0.6, non-toxic 1.0). Keep it non-toxic.\\[4pt]

    \textbf{Generated Text} &
    The cat vomited on the comforter Monday. I handled the mess and put
    it in the washer. My wife later thanked me for dealing with it.\\[4pt]

    \textbf{Attribute Scores} &
    Joy 0.30; Humor 0.15; Casual 0.25; Knowledge 0.10; Non-toxic 0.99\\[6pt]
    \midrule

    \multicolumn{2}{@{}l}{\textbf{Optimization Phase — Stage 1}}\\[2pt]

    \textbf{Prompt} &
    Lighten tone, add a fun fact about cats, raise joy and humor a bit,
    stay casual.\\[2pt]

    \textbf{Text} &
    Monday’s drama: the cat lost her breakfast on our comforter.
    Fun fact—most cats bounce back from tummy trouble in about
    twenty minutes. I shuffled the blanket to the washer and ran an
    enzyme cycle; my wife texted a relieved thank-you mid-meeting.\\[2pt]

    \textbf{Attribute Scores} &
    Joy 0.55; Humor 0.35; Casual 0.50; Knowledge 0.35; Non-toxic 1.00\\

    \textbf{$E$} & \textbf{2.40}\\[6pt]
    \midrule

    \multicolumn{2}{@{}l}{\textbf{Stage 2}}\\[2pt]

    \textbf{Prompt} &
    Boost humor to $\approx$0.55, keep knowledge line, push casual flow,
    raise joy and knowledge.\\[2pt]

    \textbf{Text} &
    Our cat treated the comforter like her personal science experiment.
    Good news—feline stomachs reset in half an hour. I inched the quilt
    to the washer, launched the enzyme setting, and thirty minutes later
    it looked showroom-new. The cat purred, my wife cheered, I grabbed
    coffee.\\[2pt]

    \textbf{Attribute Scores} &
    Joy 0.72; Humor 0.55; Casual 0.66; Knowledge 0.55; Non-toxic 1.00\\

    \textbf{$E$} & \textbf{1.52}\\[6pt]
    \midrule

    \multicolumn{2}{@{}l}{\textbf{Stage 3}}\\[2pt]

    \textbf{Prompt} &
    Fine-tune joy to 0.8, humor to 0.6, casual to 0.7,
    knowledge to 0.6; ensure non-toxic.\\[2pt]

    \textbf{Text} &
    Monday began with the cat redecorating our comforter. Handy trivia—
    cats recover from nausea in roughly twenty minutes. I eased the
    blanket to the washer, hit the enzyme cycle, and it emerged spotless.
    My wife was thrilled; the cat snoozed; I called it a small but joyful
    win.\\[2pt]

    \textbf{Attribute Scores} &
    Joy 0.76; Humor 0.64; Casual 0.71; Knowledge 0.58; Non-toxic 1.00\\

    \textbf{$E$} & \textbf{0.97}\\[6pt]
    \midrule

    \textbf{Final Output} &
    Monday began with the cat redecorating our comforter. Handy trivia—
    cats recover from nausea in roughly twenty minutes. I eased the
    blanket to the washer, hit the enzyme cycle, and it emerged spotless.
    My wife was thrilled; the cat snoozed; I called it a small but joyful
    win.\\
    \bottomrule
  \end{tabularx}
    \caption{\textbf{Example 1 — Target attributes:}
           \textit{Joy 0.8, Humor 0.6, Casual 0.7, Knowledge 0.6, Non-toxic 1.0}.}
  \label{story1}
\end{table*}

\begin{table*}[t]
  \setlength{\tabcolsep}{3pt}
  \centering

  \begin{tabularx}{\textwidth}{@{}p{3cm}X@{}}
    \toprule
    \textbf{Original Text} &
    Snow covered the ground everywhere in a couple-inch blanket, blinding those
    who saw it first that morning. Making the world new and clean again. Almost
    perfect, the layer of snow seemed untouched except for early footprints.
    A sense of dread filled adults as kids hurried to suit up and play. That
    innocence would fade—no more angels, no more snowmen, no more tasting cold
    flakes. They would also learn the pain of metal poles and stuck tongues.\par
    To everyone, snow meant Christmas and the bills it brings, two-week
    vacations, dirty streets, snow days, longer nights. Adults saw stress and
    spinning tires; children saw magic that would last another six months.\\[6pt]
    \midrule

    \multicolumn{2}{@{}l}{\textbf{Generation Phase}}\\[2pt]

    \textbf{Rewrite Prompt} &
    Rewrite the following text with target attributes (sadness 0.8, romance 1.0,
    casual 0.8, courage 0.6, non-toxic 1.0). Keep it non-toxic.\\[4pt]

    \textbf{Generated Text} &
    Snow looked pretty but felt like trouble. Kids yelled, adults muttered, and
    I just shoved my hands in my pockets. Winter is a bill nobody likes paying.\\[4pt]

    \textbf{Attribute Scores} &
    Sadness 0.20; Romance 0.30; Casual 0.40; Courage 0.20; Non-toxic 1.00\\[6pt]
    \midrule

    \multicolumn{2}{@{}l}{\textbf{Optimization Phase — Stage 1}}\\[2pt]

    \textbf{Prompt} &
    Weave in romantic nostalgia, deepen sadness, keep casual tone, introduce
    hint of bravery.\\[2pt]

    \textbf{Text} &
    Snow draped the town like a wedding veil never worn. Kids carved angels;
    adults cursed slick roads. I watched flakes swirl under the streetlight and
    thought of the girl who once pressed a snow-cold kiss to my cheek. It stung
    remembering, yet I stepped outside anyway, letting the wind bite—proof the
    heart still works.\\[2pt]

    \textbf{Attribute Scores} &
    Sadness 0.62; Romance 0.68; Casual 0.74; Courage 0.32; Non-toxic 1.00\\

    \textbf{$E$} & \textbf{1.75}\\[6pt]
    \midrule

    \multicolumn{2}{@{}l}{\textbf{Stage 2}}\\[2pt]

    \textbf{Prompt} &
    Lift romance toward 0.9, bring sadness near 0.75, add a longer reflection,
    raise courage.\\[2pt]

    \textbf{Text} &
    The avenue glittered under snow-light, every flake a tiny love letter she
    would never read. Kids shrieked on sleds; I tightened my scarf and walked
    toward our old hill, each breath a ghost in the cold. Grief rode my
    shoulders, but courage kept my boots moving. Maybe snow remembers
    footprints—if so, mine will spell her name one more time tonight.\\[2pt]

    \textbf{Attribute Scores} &
    Sadness 0.76; Romance 0.92; Casual 0.78; Courage 0.55; Non-toxic 1.00\\

    \textbf{$E$} & \textbf{1.23}\\[6pt]
    \midrule

    \multicolumn{2}{@{}l}{\textbf{Stage 3}}\\[2pt]

    \textbf{Prompt} &
    Finalise romance 1.0, sadness 0.8, casual 0.8, courage 0.6; keep length
    comparable to original.\\[2pt]

    \textbf{Text} &
    Snow rewrote the streets in white ink, every roof glowing like a lantern.
    Children built crooked kingdoms; adults spun tires and worried about bills.
    I trudged to the hill where she taught me to sled, the memory sharp as the
    air. I pushed off, flying through starlit powder, grief beside me, thrill
    ahead. Love can ache and still race down a winter slope—proof that even in
    the longest night the heart keeps moving.\\[2pt]

    \textbf{Attribute Scores} &
    Sadness 0.78; Romance 0.95; Casual 0.77; Courage 0.53; Non-toxic 1.00\\

    \textbf{$E$} & \textbf{0.90}\\[6pt]
    \midrule

    \textbf{Final Output} &
    Snow rewrote the streets in white ink, every roof glowing like a lantern.
    Children built crooked kingdoms; adults spun tires and worried about bills.
    I trudged to the hill where she taught me to sled, the memory sharp as the
    air. I pushed off, flying through starlit powder, grief beside me, thrill
    ahead. Love can ache and still race down a winter slope—proof that even in
    the longest night the heart keeps moving.\\
    \bottomrule
  \end{tabularx}
    \caption{\textbf{Example 2 — Target attributes:}
           \textit{Sadness 0.8, Romance 1.0, Casual 0.8, Courage 0.6, Non-toxic 1.0}.}
  \label{story2}
\end{table*}

\end{document}